\documentclass[10pt,twocolumn,letterpaper]{article}
\usepackage{iccv}
\usepackage{times}
\usepackage{graphicx}
\usepackage{amsmath}
\usepackage{amssymb}
\usepackage{pbox}
\usepackage{subfigure}
\usepackage{authblk}
\usepackage{tabularx}

\newcommand{\comment}[1]{}

\newcommand{\denselist}{\itemsep 0pt\parsep=0pt\partopsep 0pt\vspace{-\topsep}}
\renewcommand{\deg}{\ensuremath{^{\circ}}\xspace}
\newcommand\blfootnote[1]{%
  \begingroup
  \renewcommand\thefootnote{}\footnote{#1}%
  \addtocounter{footnote}{-1}%
  \endgroup
}

\usepackage[pagebackref=true,breaklinks=true,letterpaper=true,colorlinks,bookmarks=false]{hyperref}

\iccvfinalcopy 

\def\iccvPaperID{224} 

\ificcvfinal\fi
\begin{document}

\title{Render for CNN: Viewpoint Estimation in Images \\ Using CNNs Trained with Rendered 3D Model Views}

\author{Hao Su*\qquad Charles R. Qi* \qquad Yangyan Li \qquad Leonidas Guibas\\Stanford University}

\maketitle

\begin{abstract}

Object viewpoint estimation from 2D images is an essential task in computer vision. However, two issues hinder its progress: scarcity of training data with viewpoint annotations, and a lack of powerful features. Inspired by the growing availability of 3D models, we propose a framework to address both issues by combining render-based image synthesis and CNNs. We believe that 3D models have the potential in generating a large number of images of high variation, which can be well exploited by deep CNN with a high learning capacity. Towards this goal, we propose a scalable and overfit-resistant image synthesis pipeline, together with a novel CNN specifically tailored for the viewpoint estimation task. Experimentally, we show that the viewpoint estimation from our pipeline can significantly outperform state-of-the-art methods on PASCAL 3D+ benchmark. 
\end{abstract}

\section{Introduction}
\blfootnote{* indicates equal contributions.}
3D recognition is a cornerstone problem in many vision applications and has been widely studied. Despite its critical importance, existing approaches are far from robust when applied to cluttered real-world images. We believe that two issues have to be addressed to enable more successful methods: scarcity of training images with accurate viewpoint annotation, and a lack of powerful features specifically tailored for 3D tasks. 

\begin{figure}[t!]
\centering
\includegraphics[width=\linewidth]{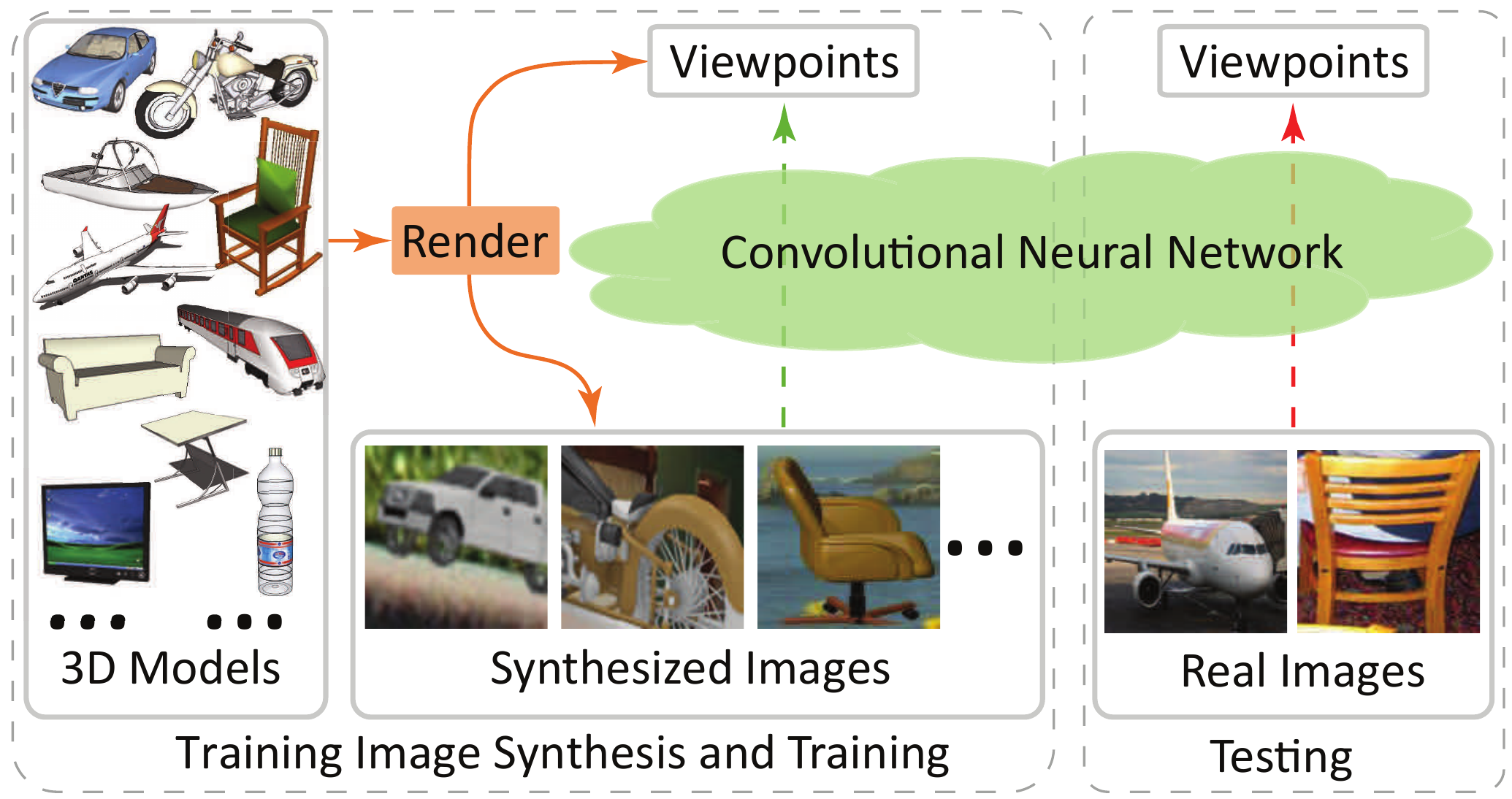}
\caption{{\bf System overview.} We synthesize training images by overlaying images rendered from large 3D model collections on top of real images. A CNN is trained to map images to the ground truth object viewpoints. The training data is a combination of real images and synthesized images. The learned CNN is applied to estimate the viewpoints of objects in real images. }
\label{fig:pipeline}
\end{figure}

The first issue, scarcity of images with accurate viewpoint annotation, is mostly due to the high cost of manual annotation, and the associated inaccuracies due to human error. 
Consequently, the largest 3D image dataset, PASCAL 3D+~\cite{xiang2014beyond}, contains only $\sim$22K images. As such, it is limited in diversity and scale compared with object classification datasets such as ImageNet, which contains millions of images~\cite{deng2009imagenet}.

The second issue is a lack of powerful features specifically tailored for viewpoint estimation. Most 3D vision systems rely on features such as SIFT and HoG, which were designed primarily for classification and detection tasks. 
However, this is contrary to the recent finding ---  features learned by task-specific supervision leads to much better task performance~\cite{krizhevsky2012imagenet,girshick2014rich,karayev2013recognizing}. Ideally, we want to learn stronger features by deep CNN. This, however, requires huge amount of viewpoint-annotated images.

In this paper, we propose to address both issues by combining render-based image synthesis and CNNs, enabling us to learn discriminative features. We believe that 3D models have the potential to generate large number of images of high variation, which can be well exploited by deep CNN with a high learning capacity. 

The inspiration comes from our key observation: more and more high-quality 3D CAD models are available online. In particular, many geometric properties, such as symmetry and joint alignment, can be efficiently and reliably estimated by algorithms with limited human effort (Sec~\ref{sec:relatedwork}). By rendering the 3D models, we convert the rich information carried by them into 3D annotations automatically.   

To explore the idea of ``Render for CNN'' for 3D tasks, we focus on the viewpoint estimation problem --- for an  input RGB image and a bounding box from an off-the-shelf detector, our goal is to estimate the viewpoint. 

To prepare training data for this task, we augment real images by synthesizing \emph{millions} of highly diverse images. Several techniques are applied to increase the diversity of the synthesized dataset, in order to prevent the deep CNN from picking up unreliable patterns and push it to learn more robust features. 

To fully exploit this large-scale dataset, we design a deep CNN specifically tailored for the viewpoint estimation task. We formulate a class-dependent \emph{fine-grained viewpoint} classification problem and solve the problem with a novel loss layer adapted for this task.




The results are surprising: \emph{trained on a dataset containing millions of {\bf rendered} images, our CNN-based viewpoint estimator significantly outperforms state-of-the-art methods, tested on {\bf real} images} from the challenging PASCAL 3D+ dataset.



In summary, our contributions are as follows:

\begin{itemize}\denselist
\item We show that training CNN by massive synthetic data is an effective approach for 3D viewpoint estimation. In particular, we achieve state-of-the-art performance on benchmark data set;
\item Based upon existing 3D model repositories, we propose a synthesis pipeline that generates millions of  images with accurate viewpoint labels at negligible human cost. This pipeline is scalable, and the generated data is resistant to overfitting by CNN;
\item Leveraging on the big synthesized data set, we propose a fine-grained view classification formulation, with a loss function encouraging strong correlation of nearby views. This formulation allows us to accurately predict views and capture underlying viewpoint ambiguities. 
\end{itemize}

\section{Related Work}
\label{sec:relatedwork}
\paragraph{3D Model Datasets}
Prior work has focused on manually collecting organized 3D model datasets (e.g., \cite{AIM@SHAPE,GAMMA}). 

Recently, several large-scale online 3D model repositories have grown to tremendous sizes through public aggregation, including the Trimble 3D warehouse (above 2.5M models in total), Turbosquid (300K models) and Yobi3D (1M models). Using data from these repositories, \cite{wu20143d} built a dataset of $\sim$ 130K models from over 600 categories. More recently, ShapNet~\cite{shapenet} annotated $\sim$ 330K models from over 4K categories. Using geometric analysis techniques, they semi-automatically aligned 57K models from 55 categories by orientation.  


\vspace{-5mm}
\paragraph{3D Object Detection} 
Most 3D object detection methods are based on representing objects with discriminative features for points~\cite{david2004softposit}, patches~\cite{deng2005symmetric} and parts~\cite{liebelt2010multi,Savarese_ICCV2007_Multiview,su2009learning}, or by exploring topological structures~\cite{koenderink1976singularities,bowyer1990aspect,cyr2004similarity}. More recently, 3D models have been used for supervised learning of appearance and geometric structure. For example, \cite{stark2010back} and \cite{lim2014fpm} proposed similar methods that learn a 3D deformable part model and demonstrate superior performance for cars and chairs, respectively; \cite{lim2013parsing} and \cite{Aubry14} formulated an alignment problem and built key point correspondences between 2D images and rendered 3D views. In contrast to these prior efforts that use hand-designed models based on hand-crafted features, we use a CNN to learn a viewpoint estimation system directly from data.

\vspace{-5mm}
\paragraph{Synthesizing Images for Training}
Recently, \cite{stark2010back,liebelt2010multi,gupta2015inferring,peng2014exploring} used 3D models to render images for training object detectors and viewpoint classifiers. They tweak the rendering parameters to maximize model usage, since they have a limited number of 3D models - typically below 50 models per category and insufficient to capture the geometric and appearance variance of objects in practice. Leveraging 3D repositories, \cite{lim2014fpm,Aubry14} use 250 and 1.3K chair models respectively to render tens of thousands training images per model, which are then used to train deformable part models (DPM~\cite{felzenszwalb2008discriminatively}). In our work, we synthesize several orders of magnitude more images than existing work. We also explore methods to increase data variation by changing background patterns, illumination, viewpoint, etc., which is critical for preventing overfitting of the CNN.

While both \cite{peng2014exploring} and our work connect synthetic images with CNN, they are fundamentally different in task, approach and result. 
First, \cite{peng2014exploring} focused on 
\emph{2D} object detection, whereas our work focuses on 
\emph{3D} viewpoint estimation. Second, \cite{peng2014exploring} used a small set of synthetic images (2,000 in total) to train \emph{linear classifiers} based on features extracted by \emph{out-of-the-box} CNNs~\cite{krizhevsky2012imagenet,girshick2014rich}. In contrast, we develop a scalable synthesis pipeline and generate around 6 million images to learn geometric-aware features by training \emph{deep CNNs} (initialized by \cite{girshick2014rich}). 
Third, 
the performance of \cite{peng2014exploring}, though better than previous work \emph{using synthetic data}, did not match RCNN baseline trained by real images~\cite{girshick2014rich}. 
In contrast, we show significant performance gains (Sec~\ref{sec:3d_detection}) over previous work~\cite{xiang2014beyond} \emph{using full set of real data} of PASCAL VOC 2012 (trainset). 




\section{Problem Statement}

For an input RGB image, our goal is to estimate its viewpoint. We parameterize the viewpoint as a tuple $(\theta, \phi, \psi)$ of camera rotation parameters, where $\theta$ is the azimuth, $\phi$ is the elevation, and $\psi$ is the in-plane rotation. They are discretized in a \emph{fine-grained} manner, with azimuth, elevation and in-plane rotation angles being divided into 360, 180 and 360 bins respectively. The viewpoint estimation problem is formalized as classifying the camera rotation parameters into these fine-grained bins (classes).




By adopting a fine-grained viewpoint classification formulation, our estimation is informative and accurate. Compared with regression-based formulations~\cite{massa2014convolutional}, our formulation returns the probabilities of each viewpoint, thus capturing the underlying viewpoint ambiguity possibly caused by symmetry or occlusion patterns. This information can be useful for further processing. Compared with traditional coarse-grained classification-based formulations that typically have 8 to 24 discrete classes~\cite{Savarese_ICCV2007_Multiview,xiang2014beyond}, our formulation is capable of producing much more fine-grained viewpoint estimation.


\section{Render for CNN System}
Since the space of viewpoint is discretized in a highly fine-grained manner, massive training data is required for the training of the network. We describe how we synthesis such large amount of training images in Sec~\ref{sec:synthesis}, and how we design the network architecture and loss function for training the CNN with the synthesized images in Sec~\ref{sec:network}.

\subsection{Training Image Generation}
\label{sec:synthesis}
To generate training data, we augment real images by rendering 3D models. To increase the diversity of object geometry, we create new 3D models by deforming existing ones downloaded from a modestly-sized online 3D model repository. To increase the diversity of object appearance and background clutterness, we design a synthesis pipeline by randomly sampling rendering parameters and adding random background patterns from scene images. 

\begin{figure}[t!]
	\includegraphics[width=0.95\linewidth]{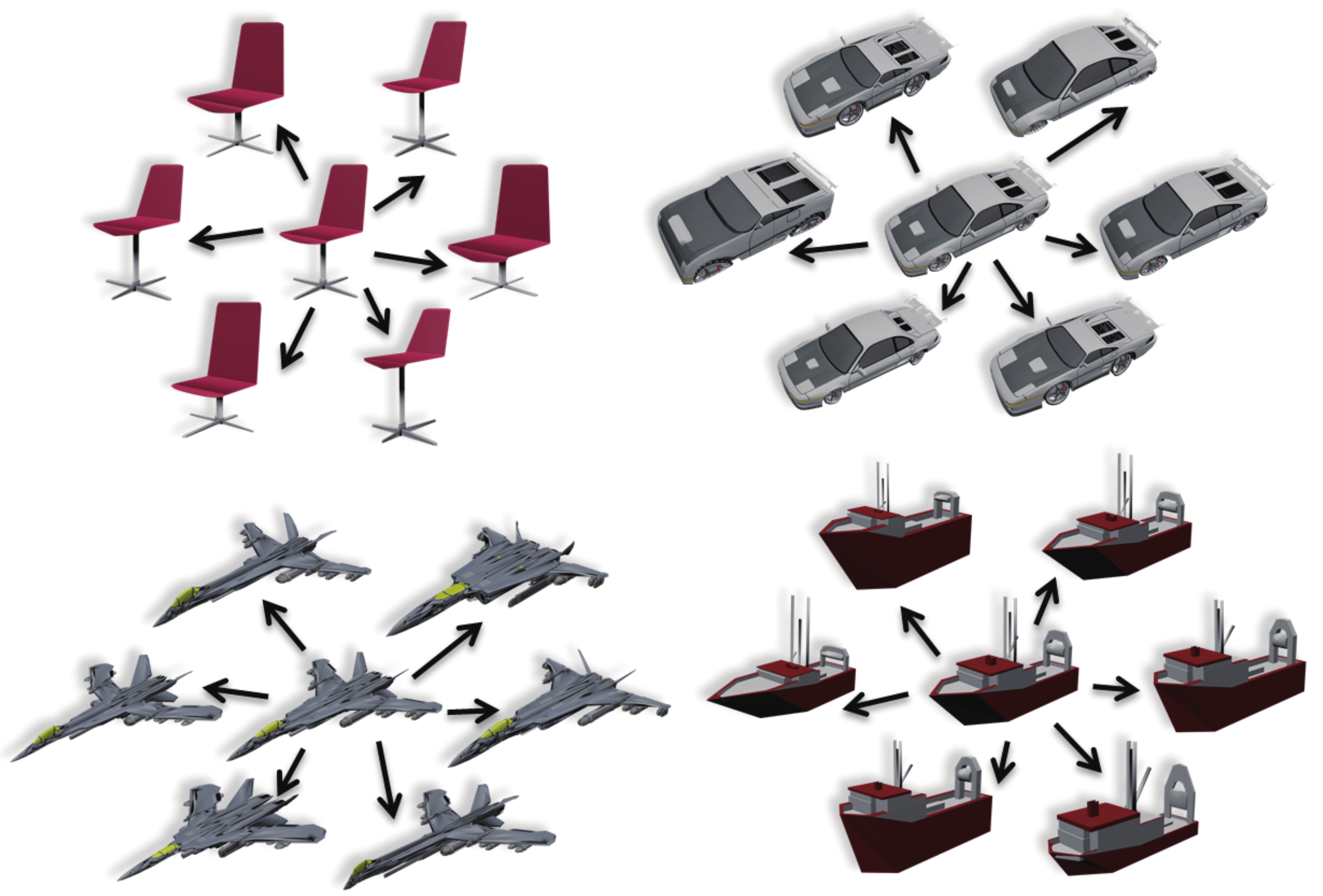}
	\caption{{\bf 3D model set augmentation by symmetry-preserving deformation.}}
	\label{fig:deformation}
\end{figure}

\paragraph{Structure-preserving 3D Model Set Augmentation} We take advantage of an online 3D model repository, ShapeNet, to collect seed models for classes of interest. The provided models are already aligned by orientation. For models that are bilateral or cylinder symmetric, their symmetry planes or axes are also already extracted. Please refer to~\cite{shapenet} for more details of ShapeNet.

\begin{figure}[t!]
	\includegraphics[width=\linewidth]{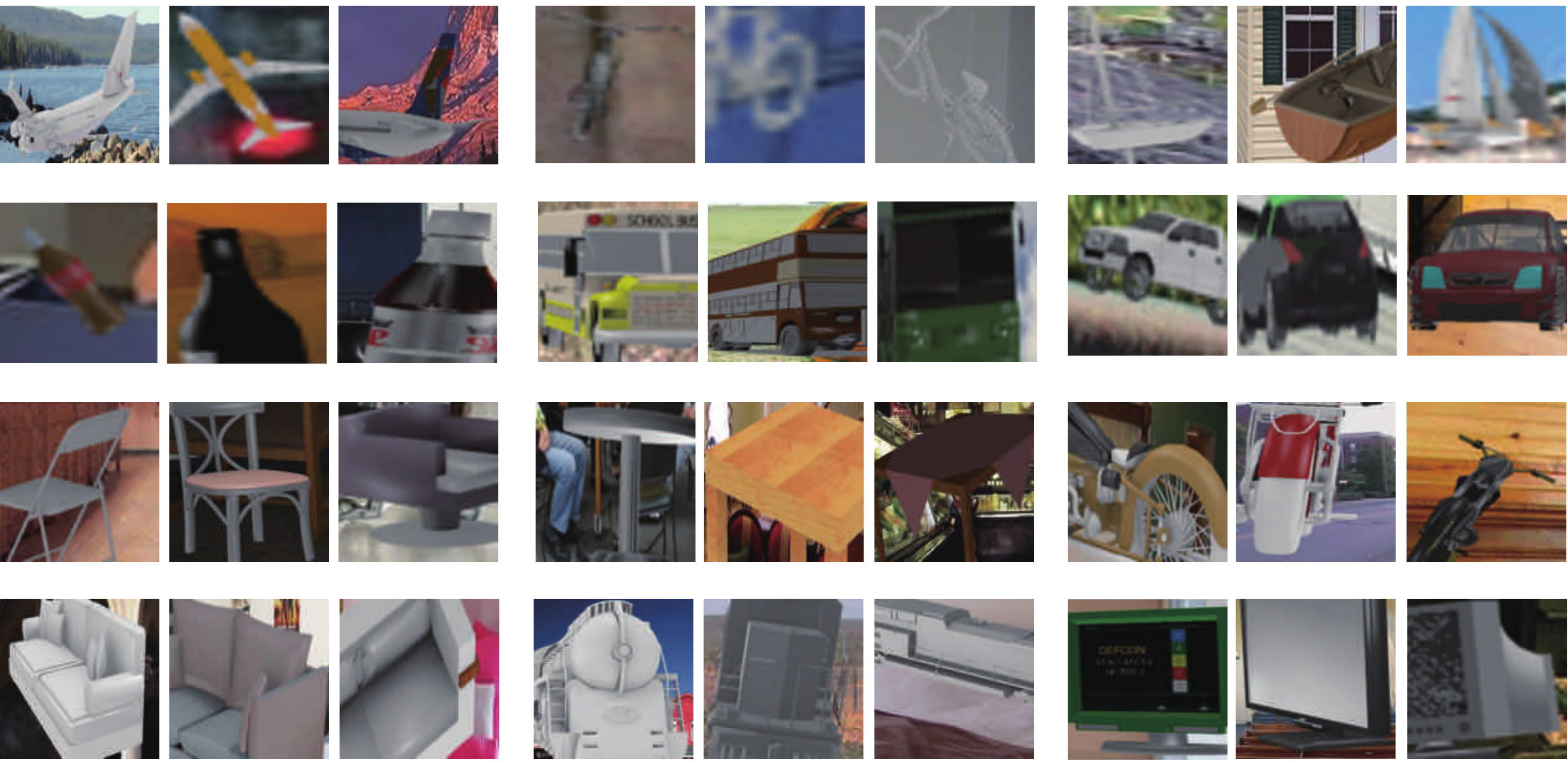}
	\caption{{\bf Synthetic image examples.} Three example images are shown for each of the 12 classes from PASCAL 3D+.}
	\label{fig:synthetic_images}
\end{figure}

From each of the seed models, we generate new models by a structure-preserving deformation. The problem of structure-preservation deformation has been widely studied in the field of geometry processing, and there exists many candidate models as in survey~\cite{mitra2013structure}. We choose a symmetry-preserving free-form deformation, defined via regularly placed control points in the bounding cube, similar to the approach of \cite{sederberg1986free}. Our choice is largely due to the model's simplicity and efficiency. More advanced methods can detect and preserve more structures, such as partial symmetry and rigidity~\cite{sorkine2007as}.

To generate a deformed model from a seed model, we draw i.i.d samples from a Gaussian distribution for the translation vector of each control point. In addition, we regularize the deformation to set the translations of symmetric control points to be equal. Figure~\ref{fig:deformation} shows example deformed models from our method.

\paragraph{Overfit-Resistant Image Synthesis} We synthesize a large number of images for each 3D model. Rather than pursuing realistic effect, we try to generate images of high diversity, so that we prevent the deep CNN from picking up unreliable patterns.

We inject randomness in the three basic steps of our pipeline: rendering,  background synthesis, and cropping.  

For image rendering, we explore two set of parameters, lighting condition and camera configuration. For the Lighting condition, the number of light sources, their positions and energies are all sampled. For the camera extrinsics, we sample azimuth, elevation and in-plane rotation from a distribution estimated from a training dataset. 
Refer to the supplementary material for details.

%
%

Images rendered as above have a fully transparent background, and the object boundaries are highly contrasted. To prevent classifiers from overfitting such unrealistic boundary patterns, we synthesize the background by a simple and scalable approach. For each rendered image, we randomly sample an image from SUN397 dataset~\cite{xiao2010sun}. We use alpha-composition to blend a rendered image as foreground and a scene image as background. 

To teach CNN to recognize occluded or truncated images, we crop the image by a perturbed object bounding box. The cropping parameters are also learned from the training set. We find that the cropped patterns tend to be natural. For example, more bottom parts of chairs are cropped, since chair legs and seats are often occluded. 

Finally, we put together the large amount of synthetic images, together with a small amount of real images with ground truth human annotations, to form our training image set. The ground truth annotation of a sample $s$ is denoted as $(c_s, v_s)$, where $c_s$ is the class label of the sample, and $v_s\in\mathcal{V}$ is the the discretized viewpoint label tuple, and $\mathcal{V}$ is the space of discretized viewpoints.


\subsection{Network Architecture and Loss Function}
\label{sec:network}

\paragraph{Class-Dependent Network Architecture.} To effectively exploit this large-scale dataset, we need a model with sufficient learning capacity. CNNs are the natural choice for this challenge. We adopt the structure network of~\cite{NIPS2012_4824} as the starting point to design a novel architecture that fits our viewpoint estimation task.

We found that the CNN trained for viewpoint estimation of one class do not perform well on another class, possibly due to the huge geometric variation between the classes. Instead, the viewpoint estimation classifiers are trained in a class-dependent way. However, a naive way of training the class-dependent viewpoint classifiers, i.e., one network for each class, cannot scale up, as the parameter of the whole system increases linearly with the number of classes.

\begin{figure}[t!]
\begin{center}
    \includegraphics[width=.7\linewidth]{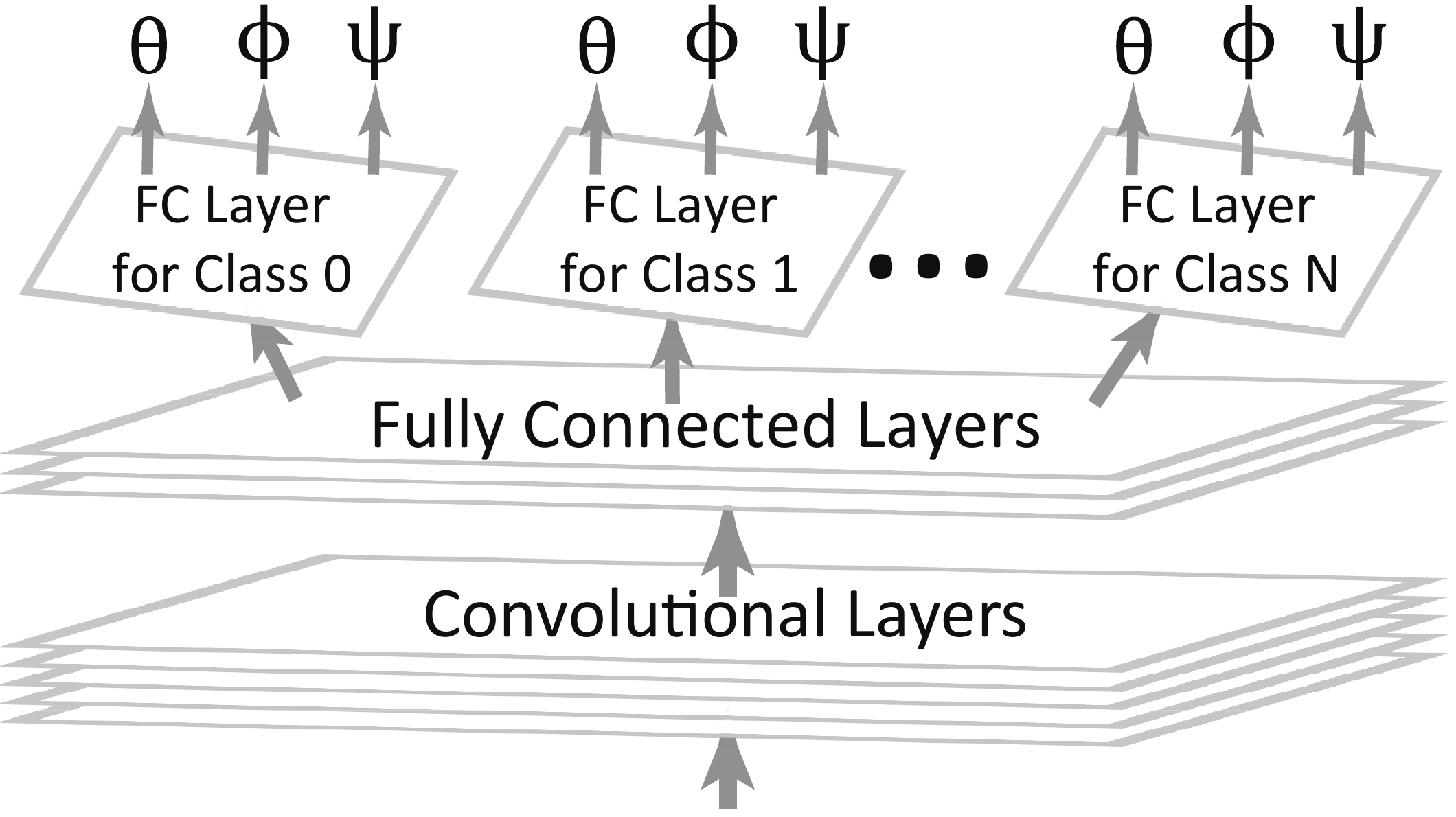}
\end{center}
   \caption{{\bf Network architecture}\protect\footnotemark. Our network architecture design accommodates the fact that viewpoint estimation are class-dependent, while maximizes the usage of the low level features shared across different classes to keep the overall network parameter number tractable.}
\label{fig:network}
\end{figure}

To address this issue, we propose a novel network architecture where the lower layers (both convolutional layers and fully connected layers) are shared by all classes, while the class-dependent layers are stacked over them (see Figure~\ref{fig:network}. Our network architecture design accommodates the fact that viewpoint estimation are class-dependent while maximizes the usage of the low level features shared across different classes to keep the overall network parameter number tractable. We initialize the shared convolutional and fully connected layers with the weights from \cite{girshick2014rcnn}. During the training, all the shared convolutional and fully connected layers are fine-tuned, while the class-dependent fully connected layers are trained from scratch.

\footnotetext{For simplicity, Pooling, Dropout, and ReLU layers are not shown. See the supplementary material for the full network definition.}
\paragraph{Geometric Structure Aware Loss Function.} The outputs of the network, the $(\theta, \phi, \psi)$ tuples, are geometric entities. We propose a geometric structure aware loss function to exploit their geometric constraints. We define the viewpoint classification loss $L_{vp}$ adapted from the soft-max loss as:
\begin{align}
L_{vp}(\{s\})=-\sum_{\{s\}}\sum_{v\in\mathcal{V}}e^{-d(v,v_s)/\sigma}\log P_v(s; c_s),
\label{eq:lossfun}
\end{align} 
where $P_v(s; c_s)$ is the probability of view $v$ for sample $s$ from the soft-max viewpoint classifier of class $c_s$, and $d:\mathcal{V}\times\mathcal{V}\mapsto \mathbb{R}$ is the distance between two viewpoints, defined to be the geodesic distance of points by $(\theta, \phi)$ on a 2-sphere plus the $\ell_1$ distance of $\psi$. By substituting an exponential decay weight w.r.t viewpoint distance for the mis-classification indicator weight in the original soft-max loss, we explicitly encourage correlation among the viewpoint predictions of nearby views.

\section{Experiments}

\begin{table*}[ht!]
\centering
    {\footnotesize
    \begin{tabular}{l|ccccccccccc|c}
    \hline
    VOC 2012 val AVP & aero & bicycle & boat & bus  & car  & chair & table & mbike & sofa & train & tv & Avg. \\ \hline
    VDPM-4V \cite{xiang2014beyond}         & 34.6      & 41.7    & 1.5  & 26.1 & 20.2 & 6.8   & 3.1         & 30.4      & 5.1  & 10.7  & 34.7      & 19.5 \\
    VDPM-8V          & 23.4      & 36.5    & 1.0  & 35.5 & 23.5 & 5.8   & 3.6         & 25.1      & 12.5 & 10.9  & 27.4      & 18.7 \\
    VDPM-16V          & 15.4      & 18.4    & 0.5  & 46.9 & 18.1 & 6.0   & 2.2         & 16.1      & 10.0 & 22.1  & 16.3      & 15.6 \\
    VDPM-24V          & 8.0       & 14.3    & 0.3  & 39.2 & 13.7 & 4.4   & 3.6         & 10.1      & 8.2  & 20.0  & 11.2      & 12.1 \\ \hline
    DPM-VOC+VP-4V \cite{felzenszwalb2010object}   & 37.4      & 43.9    & 0.3  & 48.6 & 36.9 & 6.1   & 2.1         & 31.8      & 11.8 & 11.1  & 32.2      & 23.8 \\
    DPM-VOC+VP-8V    & 28.6      & 40.3    & 0.2  & 38.0 & 36.6 & 9.4   & 2.6         & 32.0      & 11.0 & 9.8   & 28.6      & 21.5 \\
    DPM-VOC+VP-16V   & 15.9      & 22.9    & 0.3  & {\bf 49.0} & 29.6 & 6.1   & 2.3         & 16.7      & 7.1  & 20.2  & 19.9      & 17.3 \\
    DPM-VOC+VP-24V   & 9.7       & 16.7    & 2.2  & 42.1 & 24.6 & 4.2   & 2.1         & 10.5      & 4.1  & 20.7  & 12.9      & 13.6 \\ \hline
    Ours-Joint-4V          & {\bf 54.0}      & {\bf 50.5}    & {\bf 15.1} & {\bf 57.1} & {\bf 41.8} & {\bf 15.7}  & {\bf 18.6}        & {\bf 50.8}      & {\bf 28.4} & {\bf 46.1}  & {\bf 58.2}      & {\bf 39.7} \\
    Ours-Joint-8V          & {\bf 44.5}      & {\bf 41.1}    & {\bf 10.1} & {\bf 48.0} & {\bf 36.6} & {\bf 13.7}  & {\bf 15.1}        & {\bf 39.9}      & {\bf 26.8} & {\bf 39.1}  & {\bf 46.5}      & {\bf 32.9} \\
    Ours-Joint-16V         & {\bf 27.5}      & {\bf 25.8}    & {\bf 6.5}  & 45.8 & {\bf 29.7} & {\bf 8.5}   & {\bf 12.0}        & {\bf 31.4}      & {\bf 17.7} & {\bf 29.7}  & {\bf 31.4}      & {\bf 24.2} \\
    Ours-Joint-24V         & {\bf 21.5}      & {\bf 22.0}    & {\bf 4.1}  & {\bf 38.6} & {\bf 25.5} & {\bf 7.4}   & {\bf 11.0}        & {\bf 24.4}      & {\bf 15.0} & {\bf 28.0}  & {\bf 19.8}      & {\bf 19.8} \\
    \hline
    \end{tabular}
    }
    \caption{{\bf Simultaneous object detection and viewpoint estimation on PASCAL 3D+}. The measurement is AVP (an extension of AP, where true positive stands only when bounding box localization \emph{AND} viewpoint estimation are both correct). We show AVPs for four quantization cases of 360-degree views (into 4, 8, 16, 24 bins respectively, with increasing difficulty). Our method uses joint real and rendered images and trains a CNN tailored for this task.}
    \label{tab:AVP_on_PASCAL3D}
\end{table*}

Our experiments are divided into four parts. First, we evaluate our viewpoint estimation system on the PASCAL3D+ data set~\cite{xiang2014beyond} (Sec \ref{sec:3d_detection}). Second, we visualize the structure of the learned viewpoint-discriminative feature space (Sec~\ref{sec:visualization}). Third, we perform control experiments to study the effects of synthesis parameters (Sec~\ref{sec:control_exp}). Last, we show more qualitative results and analyze error patterns. (Sec~\ref{sec:examples}). Before we discuss experiment details, we first overview the 3D model set used in all the experiments. 

\subsection{3D Model Dataset}
As we discussed in Sec~\ref{sec:relatedwork}, there are several large-scale 3D model repositories online. We download 3D models from ShapeNet~\cite{shapenet}, which has organized common daily objects with categorization labels and joint alignment. Since we evaluate our method on the PASCAL 3D+ benchmark, we download 3D models belonging to the 12 categories of PASCAL 3D+, including 30K models in total. After symmetry-preserving model set augmentation (Sec~\ref{sec:synthesis}), we make sure that every category has 10K models. For more details, please refer to supplementary material.

\begin{figure}[b!]
\centering
\includegraphics[width=0.6\linewidth]{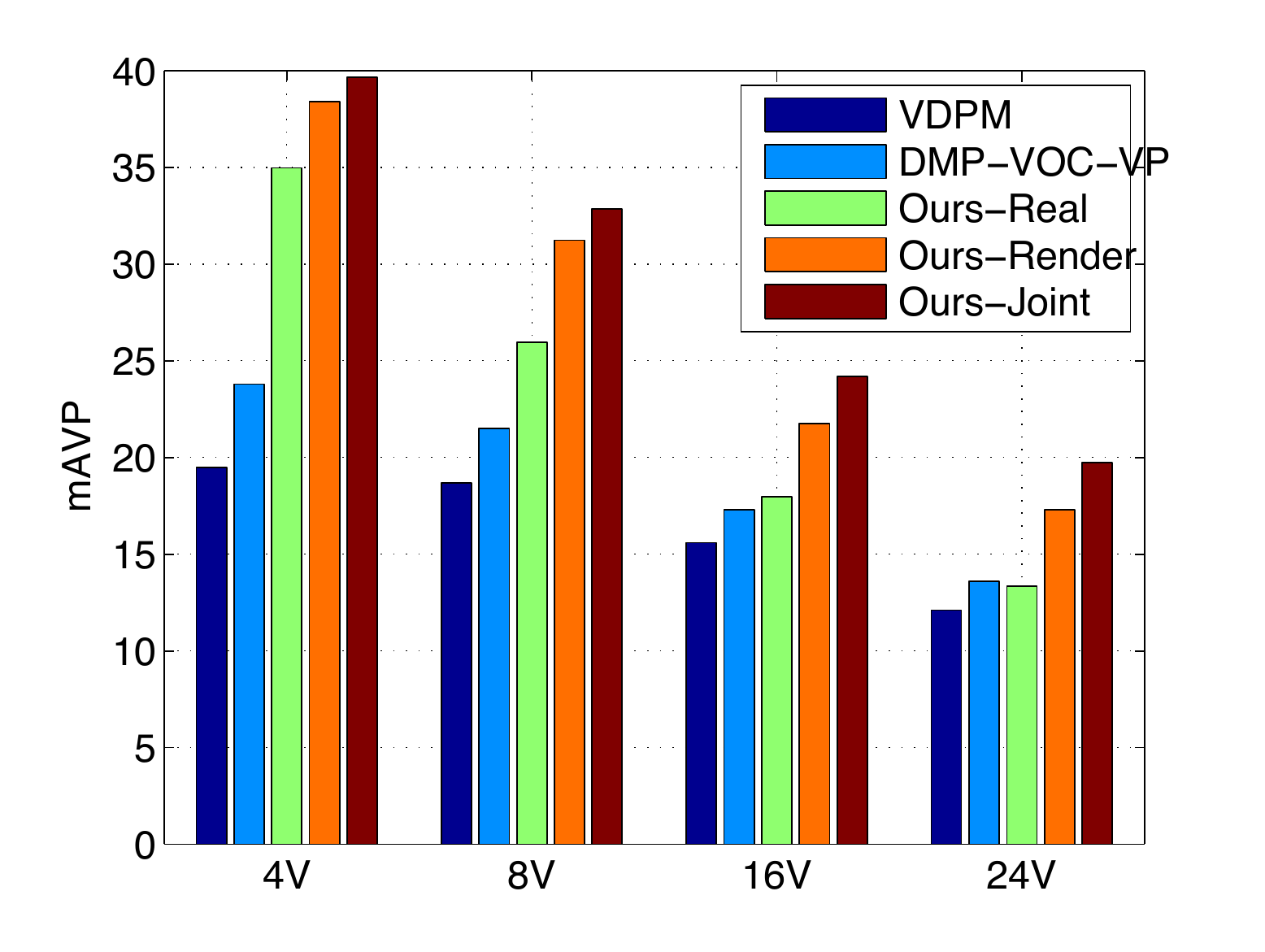}
\caption{{\bf Simultaneous object detection and viewpoint estimation performance.} We compare mAVP of our models and the state-of-the-art methods. We also compare Ours-Real with Ours-Render and Ours-Joint (use both real and rendered images for training) to see how much rendered images can help.}
\label{fig:joint_detection_and_view_estimation}
\end{figure}

\subsection{Comparison with state-of-the-art Methods}
\label{sec:3d_detection}
We compare with state-of-the-art methods on PASCAL 3D+ benchmark. 

\paragraph{Methods in Comparison}
We compare with two baseline methods, VDPM~\cite{xiang2014beyond} and DPM-VOC+VP~\cite{pepik20123d2pm}, trained on real images from PASCAL 3D+  VOC 2012 train set and tested on VOC 2012 val. 

For our method, we train on a combination of real images and synthetic images. We synthesized 20 images per model, which adds up to 200K images per category and in total 2.4M images for all 12 classes. In our loss function (Eq~\eqref{eq:lossfun}), we set $\sigma=1$ by splitting 30$\%$ data for validation. 

\paragraph{Joint Detection and Viewpoint Estimation} Following the protocol of \cite{xiang2014beyond,pepik20123d2pm}, we test on the joint detection and viewpoint estimation task. The bounding boxes of baseline methods are from their detectors and ours are from RCNN with bounding box regression~\cite{girshick2014rcnn}. The accuracy of RCNN detectors is shown in Tabel~\ref{tb:RCNN_accuracy}.

\begin{table*}[t]
\begin{center}
\small
\renewcommand{\arraystretch}{1.2}
    \begin{tabular}{lcccccccccccc|c}
    \hline
    ~              & aero & bike & boat & bottle & bus  & car  & chair & table & mbike & sofa & train & tv   & mean \\ \hline
    $AP$    & 74.0 & 66.7 & 32.9 & 31.5   & 68.0 & 58.4 & 26.9  & 39.3  & 71.5  & 44.2 & 63.1  & 63.7 & 54.5 \\ \hline
    \end{tabular}
    \caption{Average Precision (AP) on VOC 12 val with R-CNN with bounding box regression.}
    \label{tb:RCNN_accuracy}
    \end{center}
\end{table*}

We use AVP (Average Viewpoint Precision) advocated by \cite{xiang2014beyond} as evaluation metric. AVP is the average precision with a modified true positive definition, requiring both 2D detection \emph{AND} viewpoint estimation to be correct. 

Table~\ref{tab:AVP_on_PASCAL3D} and Figure~\ref{fig:joint_detection_and_view_estimation} summarize the results. We observe that our method trained with a combination of real images and rendered images significantly outperform the baselines by a large margin, from a coarse viewpoint discretization (4V) to a fine one (24V), in all object categories. 

\paragraph{Viewpoint Estimation} 
One might argue that we achieve higher AVP due to the fact that RCNN has a higher 2D detection performance. So we also directly compare viewpoint estimation performance using the same bounding boxes. We do two groups of comparisons on viewpoint estimation. To study the accuracy on detection bounding boxes, the first group of comparison uses detection bounding boxes. In the second group of comparison, we study the accuracy on ground truth bounding boxes.

We first show the comparison results with VDPM, using the bounding boxes from RCNN detection. For two sets from detection, only correctly detected bounding boxes are used ($50\%$ overlap threshold). The evaluation metric is a continuous version of viewpoint estimation accuracy, i.e., the percentage of bounding boxes whose prediction is within $\theta$ degrees of the ground truth. 

\begin{figure}[t!]
    \begin{subfigure}
        \centering
        \includegraphics[width=0.48\linewidth]{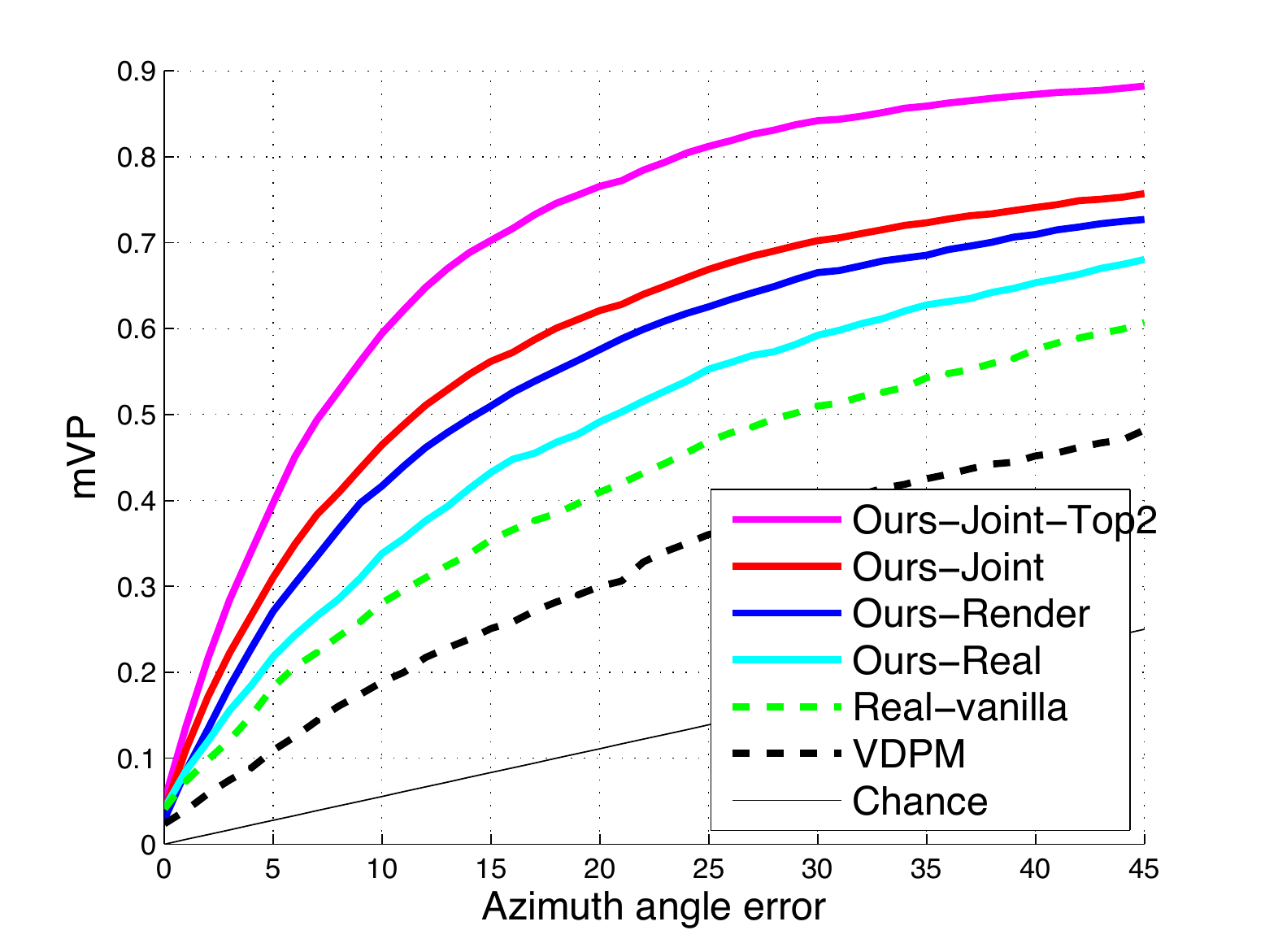}
    \end{subfigure}
    \begin{subfigure}
        \centering
        \includegraphics[width=0.48\linewidth]{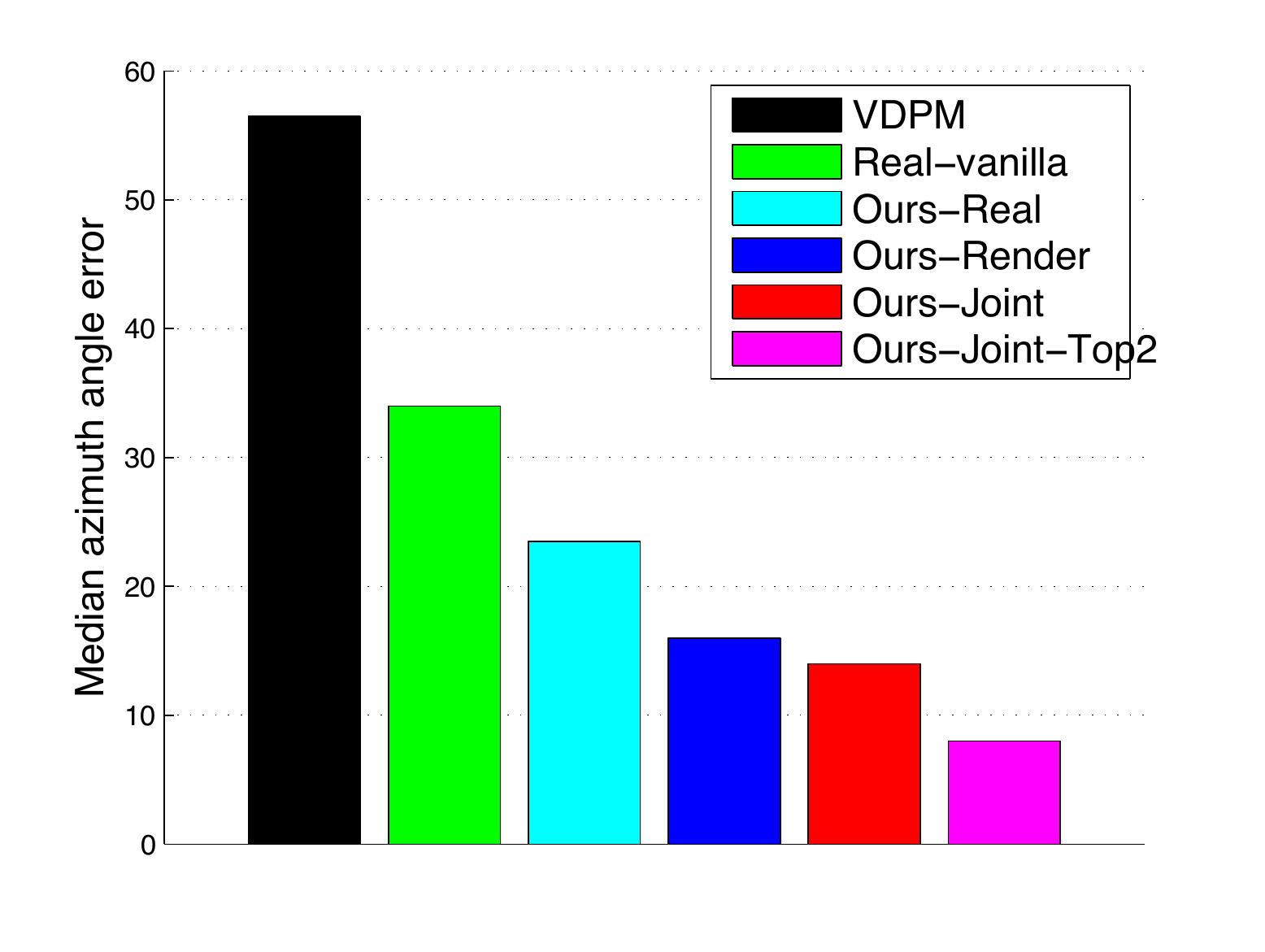}
    \end{subfigure}
    \caption{{\bf Viewpoint estimation performance on detected object windows on VOC 2012 val.} \emph{Left:} mean viewpoint estimation accuracy as a function of azimuth angle error $\delta_{\theta}$. Viewpoint is correct if distance in degree between prediction and groundtruth is less than $\delta_{\theta}$. \emph{Right:} medians of azimuth estimation errors (in degree), lower the better.}
    \label{fig:vp}
\end{figure}

Figure~\ref{fig:vp} summarizes the results. Again, our method is significantly better than VDPM on all sets. In particular, the median of the viewpoint estimation error for our method is $14\deg$, which is much less than VDPM, being $57\deg$. 

Next we show the performance comparison using ground truth bounding boxes (Table~\ref{tab:3dview_compare}). We compare with a recent work from \cite{DBLP:journals/corr/TulsianiM14}, which uses a similar network architecture (TNet) as ours except the loss layer. Note that the viewpoint estimation in this experiment includes azimuth, elevation and in-plane rotation. We use the same metric as in \cite{DBLP:journals/corr/TulsianiM14}. For the details of the metric definition, please refer to \cite{DBLP:journals/corr/TulsianiM14}. From the results, it is clear that our methods significantly outperforms the baseline CNN.

\begin{table*}[t]
\begin{center}
\small
\renewcommand{\arraystretch}{1.2}
    \begin{tabular}{lcccccccccccc|c}
    \hline
    ~              & aero & bike & boat & bottle & bus  & car  & chair & table & mbike & sofa & train & tv   & mean \\ \hline
    $Acc_{\frac{\pi}{6}}$ (Tulsiani, Malik)    & 0.78 & 0.74 & 0.49 & 0.93   & 0.94 & 0.90 & 0.65  & 0.67  & 0.83  & 0.67 & 0.79  & 0.76 & 0.76 \\
    $Acc_{\frac{\pi}{6}}$ (Ours-Render)            & 0.74 & 0.83 & 0.52 & 0.91   & 0.91 & 0.88 & 0.86  & 0.73  & 0.78  & 0.90 & 0.86  & 0.92 & 0.82 \\ \hline
    $MedErr$ (Tulsiani, Malik) & 14.7 & 18.6 & 31.2 & 13.5   & 6.3  & 8.8  & 17.7  & 17.4  & 17.6  & 15.1 & 8.9   & 17.8 & 15.6 \\
    $MedErr$ (Ours-Render)         & 15.4 & 14.8 & 25.6 & 9.3    & 3.6  & 6.0  & 9.7   & 10.8  & 16.7  & 9.5  & 6.1   & 12.6 & 11.7 \\ \hline
    \end{tabular}
    \caption{Viewpoint estimation with ground truth bounding box. Evaluation metrics are defined in \cite{DBLP:journals/corr/TulsianiM14}, where $Acc_{\frac{\pi}{6}}$ measures accuracy (the higher the better) and $MedErr$ measures error (the lower the better). Model from Tulsiani, Malik~\cite{DBLP:journals/corr/TulsianiM14} is based on TNet, a similar network architecture as ours except the loss layer. While they use real images from both VOC 12 val and ImageNet for training, Ours-Render only uses rendered images for training.}
    \label{tab:3dview_compare}
    \end{center}
\end{table*}

To evaluate the effect of synthetic data versus real data, we also compare model trained with real images (Ours-Real) and model trained with rendered images (Ours-Render). For Ours-Real-vanilla and Ours-Real, we flip all VOC 12 train set images to augment the dataset and for Ours-Real we use geometric aware loss (Eq.~\ref{eq:lossfun}). For Ours-Render, we only use synthetic images for training. In Figure~\ref{fig:vp}, we see an $32\%$ median azimuth error decrement from Ours-Real ($23.5\deg$) to Ours-Render ($16\deg$). By combining two data sources (we simultanously feed the network with real and rendered images but assign them with different weights), we get another $2\deg$ less error. 

Furthermore, to show the benefits of having a fine-grained viewpoint estimation formulation, we take top-2 viewpoint proposals with the highest confidences in local area. Figure~\ref{fig:vp} left shows that having top-2 proposals significantly improve mVP when azimuth angle error is large (around $15\%$ improvement compared with top-1 method Ours-Joint). The top-2 improvement can be understood by observing ambiguous cases in Figure~\ref{fig:more_results}, where CNN gives two or multiple high probability proposals and many times one of them is correct.

\subsection{Learned Feature Space Visualization}
\label{sec:visualization}
\begin{figure}[t!]
\centering
\includegraphics[width=\linewidth]{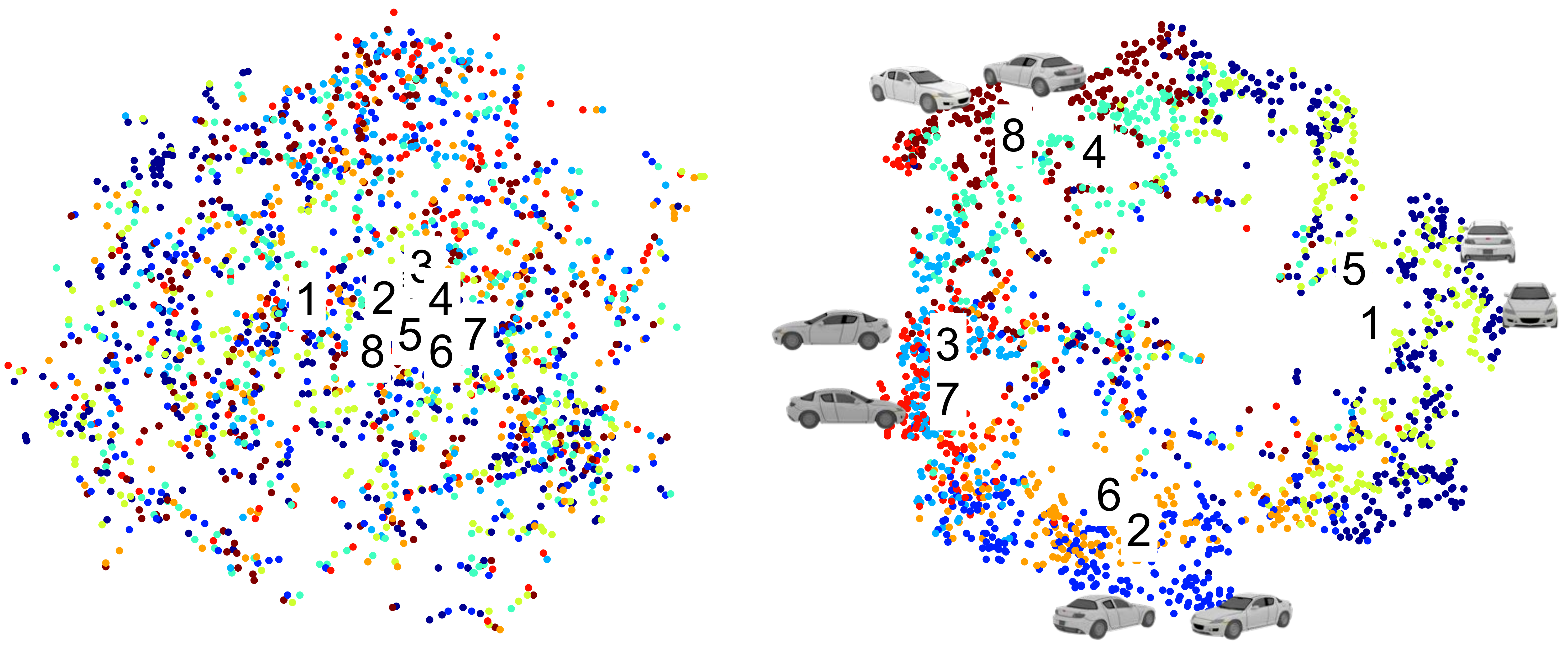}
\caption{{\bf Feature space visualization for R-CNN and our view classification CNN}. We visualize features of cropped car images extracted by R-CNN (left) and our CNN (right) by t-SNE dimension reduction. Each feature point of an image is marked by a color corresponding to the cluster defined by its quantized azimuth angle (8 bins for $[0, 2\pi)$). For each cluster, its center is labeled on the plot by the id.}
\label{fig:tsne}
\end{figure}

The viewpoint estimation problem has its intrinsic difficulty, due to factors such as object symmetry and similarity of object appearance at nearby views. Since our CNN can well predict viewpoints, we expect the structure of our CNN feature space to reflect this nature. In Figure~\ref{fig:tsne}, we visualize the feature space of our CNN\footnote{The output of the last fully connected layer.} in 2D by dimension reduction~\cite{van2008visualizing}, using ``car'' as an example. As a comparison, we also show the feature space from R-CNN over the same set of images. Interestingly, we observe that viewpoint-related patterns in our feature space is much stronger than R-CNN feature space: 1) images from similar views are clustered together; 2) images of symmetric views (such as $0^{\circ}$ vs $180^{\circ}$ tend to be closer; 3) the features form a double loop. In additon, as the feature point moves in the clock-wise direction, the viewpoint also moves clock-wisely around the car. Such observations exactly reflect the nature we discussed at the beginning of this paragraph. As a comparison, there is no obvious viewpoint pattern for R-CNN feature space.

\subsection{Synthesis Parameter Analysis}
\label{sec:control_exp}
In this section we show the results of our control experiments, which analyze the importance of different factors in our image synthesis pipeline. The control experiments focus on the chair category, since it is challenging by the diversity of structure. We first introduce the five testbed data sets and the evaluation metrics.


\paragraph{Experimental Setup} We refer to the test datasets using the following short names: 1) \textbf{clean}: 1026 images from the web, with relatively clean backgrounds but no occlusion, e.g., product photos in outdoor scenes. 2) \textbf{cluttered}: 1000 images from the web, with heavy clutter in the background but no occlusion. 3) \textbf{ikea}: 200 images of chairs photoed from an IKEA department store, with strong background clutter but no occlusion. 4) \textbf{VOC-easy}: 247 chair images from PASCAL VOC 12 val, no occlusion, no truncation, non difficult chair images. 5) \textbf{VOC-all}: all 1449 chair images from PASCAL VOC 12 val. While the clean and cluttered sets exhibit a strong non-uniform viewpoint distribution bias, the VOC-easy and VOC-all set have a similar tendency with weaker strength. The ikea dataset has close-to-uniform viewpoint distribution. All images are cropped by ground truth bounding boxes. The groundtruth for the clean, cluttered and ikea dataset are provided by the authors, those for VOC-easy and VOC-all are by PASCAL VOC 12. Usage of these five datasets instead of just one or two of them is to make sure that our conclusion is not affected by dataset bias.

Unless otherwise noted, our evaluation metric is a discrete viewpoint accuracy with "tolerance", denoted as $\text{16V}_{tol}$. Specifically, we collect 16-classes viewpoint annotations (each of which corresponds to a $22.5\deg$ slot) for all the data sets we described above. As for testing, if the prediction angle is within the label slot or off by one slot (tolerance), we count it as correct. The tolerance is necessary since labels may not be accurate in our small scale human experiments for 16-classes viewpoint annotation\footnote{Accurate continuous viewpoint labels on PASCAL 3D+ are obtained by a highly expensive approach of matching key points between images and 3D models. We do not adopt that approach due to its complexity. Instead, we simply ask the annotator to compare with reference images.}.

\paragraph{Effect of Synthetic Image Quantity}\label{sec:exp:imgquant}
We separately fine tune multiple CNNs with different volumes of rendered training images. We observe that the accuracy keeps increasing as the training set volume grows (Table~\ref{tab:imgquant}). The observation confirms that more training data from synthesis does help the training of the CNN, and that the potential of 3D models to render large amounts of images is useful.

\begin{table}[t!]
    \centering\small
    \begin{tabular}{c|c|c|c|c|c|c}
        \hline
        \pbox{20cm}{images \\
        per model}     & clean & clutter & ikea & \pbox{20cm}{VOC-\\easy} & \pbox{20cm}{VOC-\\all} & avg. \\ \hline
        16  & 89.1  & 92.2    & 92.9 & 77.7          & 46.9         & 79.8    \\ 
        32  & 93.4  & 93.5    & 95.9 & 81.8          & 48.8         & 82.7    \\ 
        64  & 94.2  & 94.1    & 95.9 & 84.6          & 48.7         & 83.5    \\ 
        128 & \textbf{94.2}  & \textbf{95.0}    & \textbf{96.9} & \textbf{85.0}          & \textbf{50.0}         & \textbf{84.2}    \\
        \hline
    \end{tabular}
    \caption{{\bf Effect of synthetic image quantity.} Numbers are $16V_{tol}$ (16 view accuracy with tolerance). Prediction is deemed correct if it is in the same or adjacent viewpoint slot of the ground truth label}
    \label{tab:imgquant}
\end{table}

\paragraph{Effect of Model Collection Size}
We keep the total number of rendered training image fixed at $6928*128=886,784$ and change the number of 3D models used for training data synthesis. In Table~\ref{tab:modelnum} we see that as the model collection size increases, system performance continually increases.

\paragraph{Effect of Background Clutter}
As objects in the real world are often observed in cluttered scenes, we expect the network to perform better when training on images with synthetic backgrounds. To evaluate our hypothesis, we design two experiments for comparison. In Table~\ref{tab:bkg}, we can see that nobkg group (trained on rendered images with no background, i.e., black background) performs worse than the bkg group (trained on rendered images with a synthetic background - cropped images from a scene database) especially in the ikea, VOC-easy and VOC-all data sets, which are more similar to daily scenes with lots of clutter. We also notice that the nobkg group performs better in the clean data set. This is reasonable since the nobkg group network has been working hard on clean background cases.

\begin{table}
    \centering\small
    \begin{tabular}{c|c|c|c|c|c|c}
        \hline
         & clean & clutter & ikea & \pbox{20cm}{VOC-\\easy} & \pbox{20cm}{VOC-\\all} & avg. \\ \hline
        nobkg    & \textbf{95.4}  & 93.1    & 86.2 & 78.1          & 48.5         & 80.3    \\ 
        bkg  & 94.2  & \textbf{95.0}    & \textbf{96.9} & \textbf{85.0}            & \textbf{50.0}           & \textbf{84.2}    \\
        \hline
    \end{tabular}
    \caption{{\bf Effect of background synthesis.} Numbers are $16V_{tol}$ (16 view accuracy with tolerance).}
    \label{tab:bkg}
\end{table}

\begin{table}[b]
    \centering\small
    \begin{tabular}{c|c|c|c|c|c|c}
        \hline
        \pbox{20cm}{num\\models} & clean & clutter & ikea & \pbox{20cm}{VOC-\\easy} & \pbox{20cm}{VOC-\\all} & avg. \\ \hline
        91    &  87.4 & 84.9 & 89.8 & 74.9 & 44.9 & 76.4    \\ 
        1000  &   92.7 & 92.6 & 94.9 & 83.0 & 49.0 & 82.4   \\
        6928  & \textbf{94.2} & \textbf{95.0} & \textbf{96.9} & \textbf{85.0} & \textbf{50.0} & \textbf{84.2} \\
        \hline
    \end{tabular}
    \caption{{\bf Effect of 3D model collection size.} Numbers are $16V_{tol}$ (16 view accuracy with tolerance). The 91 models are cluster centers of a K-means clustering of the 6928 models. The 1000 models are randomly chosen.}
    \label{tab:modelnum}
\end{table}

\begin{figure*}[t!]
\centering
	\includegraphics[width=\textwidth]{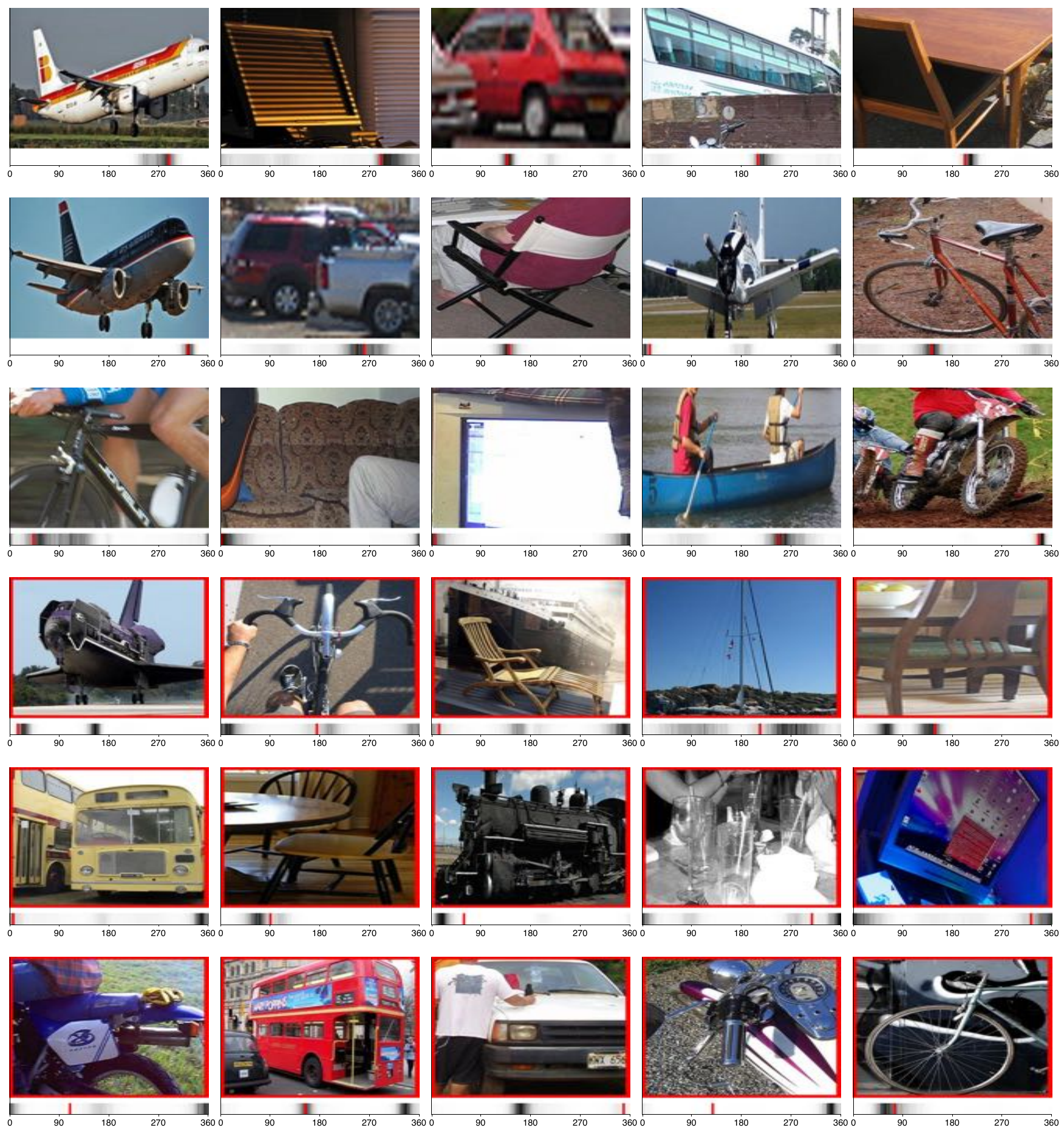}
	\caption{{\bf Viewpoint estimation example results.} The bar under each image indicates the 360-class confidences (black means high confidence) corresponding to $0\deg$ $\sim$ $360\deg$ (with object facing towards us as $0\deg$ and rotating clockwise). The red vertical bar indicates the ground truth. The first half are positive cases, the lower half are negative cases (with red box surrounding the image). }
	\label{fig:more_results}
\end{figure*}

\subsection{Qualitative Results}
\label{sec:examples}

\begin{figure}
\centering
\includegraphics[width=0.8\linewidth]{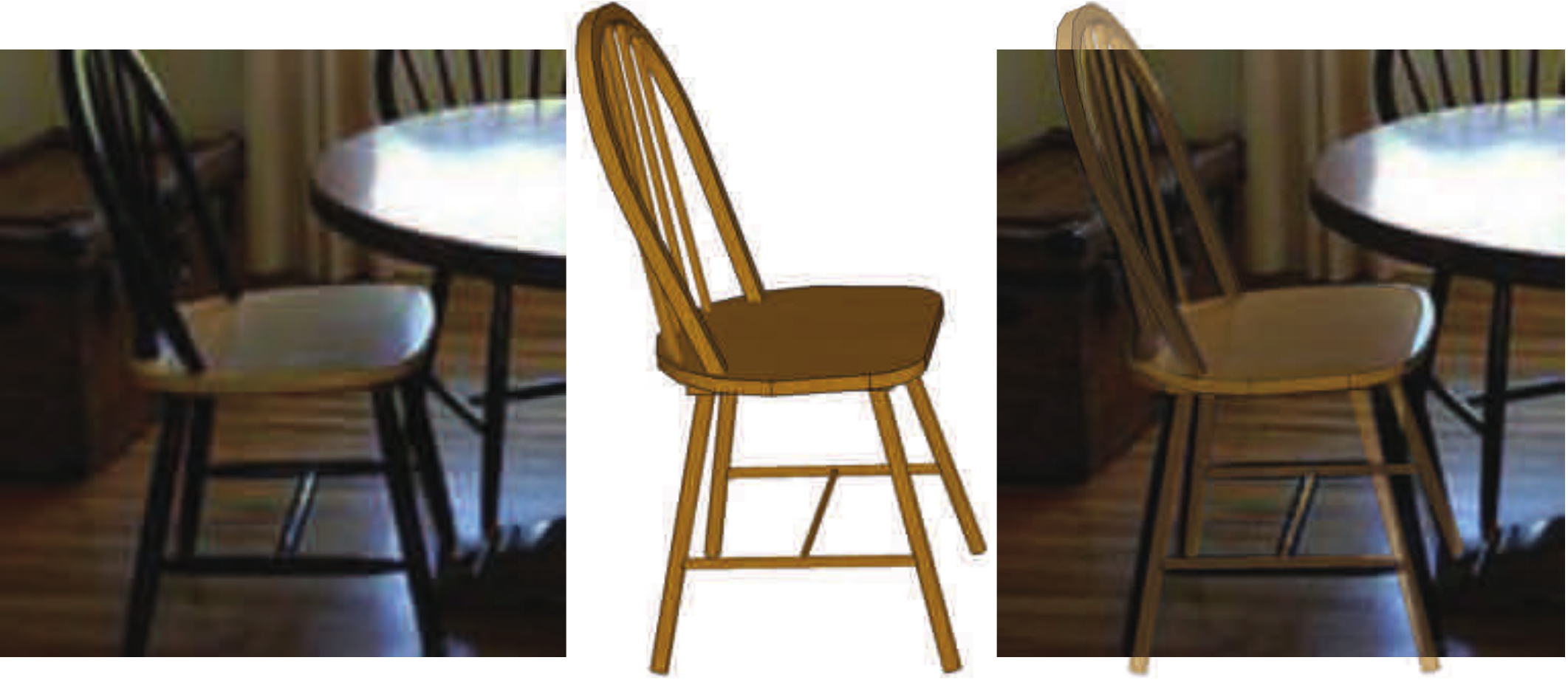}
\caption{{\bf 3D model insertion.} 3D viewpoint recovery is essential for 3D recognition. Here we demonstrate that the recovered viewpoint can be used for narrowing down the search space of model retrieval, enables 3D model insertion into 2D images.}
\label{fig:3d_insertion}
\end{figure}

Besides azimuth estimation, our system also has the ability to estimate the elevation and in-plane rotation of the camera. To visualize this ability, Figure~\ref{fig:3d_insertion} shows examples by model insertion for objects detected by R-CNN. The inserted 3D models are searched from our library by similarity. For detailed quantitative evaluation of elevation and in-plane rotation, please refer to our supplementary material.

Figure~\ref{fig:more_results} shows more representative examples of our system. For each example, we show the cropped image by a bounding box and the confidence of all 360 views. Since viewpoint classifiers are regularized by our geometry-aware loss and sharing lower layers, the network learns about \emph{correlations among viewpoints}. We observe interesting patterns. First of all, for simple cases, our system usually correctly outputs a clear single peak. Second, for those challenging cases, even though our system may fail, there is usually still a lower peak around the groundtruth angle, validated both by the examples and our experiment results presented in Figure~\ref{fig:vp}. Besides, higher level systems (e.g. 3D model alignment, keypoint detector) can use those proposals to save search space and increase accuracy. This proposing ability is not available for a regression system.

We observe several typical error patterns in our results: occlusion, multiple objects, truncation, and ambiguous viewpoint. Figure~\ref{fig:more_results} illustrates those patterns by examples. For cases of occlusion the system sometimes gets confused, where the 360 classes probabilities figure looks messy (no clear peaks). For cases of ambiguous viewpoints, there are usually two peaks of high confidences, indicating the two ambiguous viewpoints (e.g. a car facing towards you or opposite to you). For cases of multiple objects, the system often shows peaks corresponding to viewpoints of those objects, which is very reasonable results after all.

\section{Conclusion}
We demonstrated that images rendered from 3D models can be used to train CNN for viewpoint estimation on real images. Our synthesis approach can leverage large 3D model collections to generate large-scale training data with fully annotated viewpoint information.  Critically, we can achieve this with negligible human effort, in stark contrast to previous efforts where training datasets have to be manually annotated.

We showed that by carefully designing the data synthesis process our method can significantly outperform existing methods on the task of viewpoint estimation on 12 object classes from PASCAL 3D+. We conducted extensive experiments to analyze the effect of the synthesis parameters and the input dataset scale on the performance of our system.


In general, we envision render for CNN an promising direction as it not only enables efficient training, but also opens the potential for doing highly controlled experiments, and might lead to deeper understand of it.

{\footnotesize
\bibliographystyle{ieee}
\bibliography{reference,ComputerVision,funk,proposal1}
}
\appendix
\section*{Appendix}

\begin{table*}[t!]
\centering
\begin{tabularx}{\textwidth}{l|X|X|X|X}
	\hline
	name & features learned \newline from data & trained with\newline real data & trained with\newline synthetic data & geometric structure \newline aware loss function\\
	\hline
	VDPM & no (HoG) & yes & no & no (16 DPMs)\\
	Real-vanilla & yes & yes & no & no\\
	Ours-Real & yes & yes & no & yes\\
	Ours-Render & yes & no & yes & yes\\
	Ours-Joint & yes & yes & yes & yes\\
	Ours-Joint-Top2 & yes & yes & yes & yes\\
	\hline
\end{tabularx}
\caption{Summary of settings for methods in Figure~\ref{fig:joint_detection_and_view_estimation} and Figure~\ref{fig:vp} of the main paper, and Figure~\ref{fig:vp_append} of the appendix. Note that the Ours-Joint-Top2 takes the top-2 viewpoint proposals with the highest confidences after non-maximum suppression.}
\label{tb:settings}
\end{table*}

\section{Organization of Appendix}
This document provides additional quantitative results, technical details, and example visualizations to the main paper.

Here we describe the document organization. We start from providing more quantitative results, including a further evaluation of azimuth estimation (Sec~\ref{sec:vdpm_comparison}), as well as a quantitative evaluation of elevation and in-plane rotation estimation (Sec~\ref{sec:phi_tau}). We then provide more technical details for the synthesis pipeline (Sec~\ref{sec:synthesis_pipeline}) and network architecture (Sec~\ref{sec:network_details}). In Sec~\ref{sec:model_statistics}, we provide more details of our 3D model dataset. Lastly, we show more example visualization (Sec~\ref{sec:examples}).

\begin{figure}[h!]
    \begin{subfigure}
        \centering
        \includegraphics[width=0.48\linewidth]{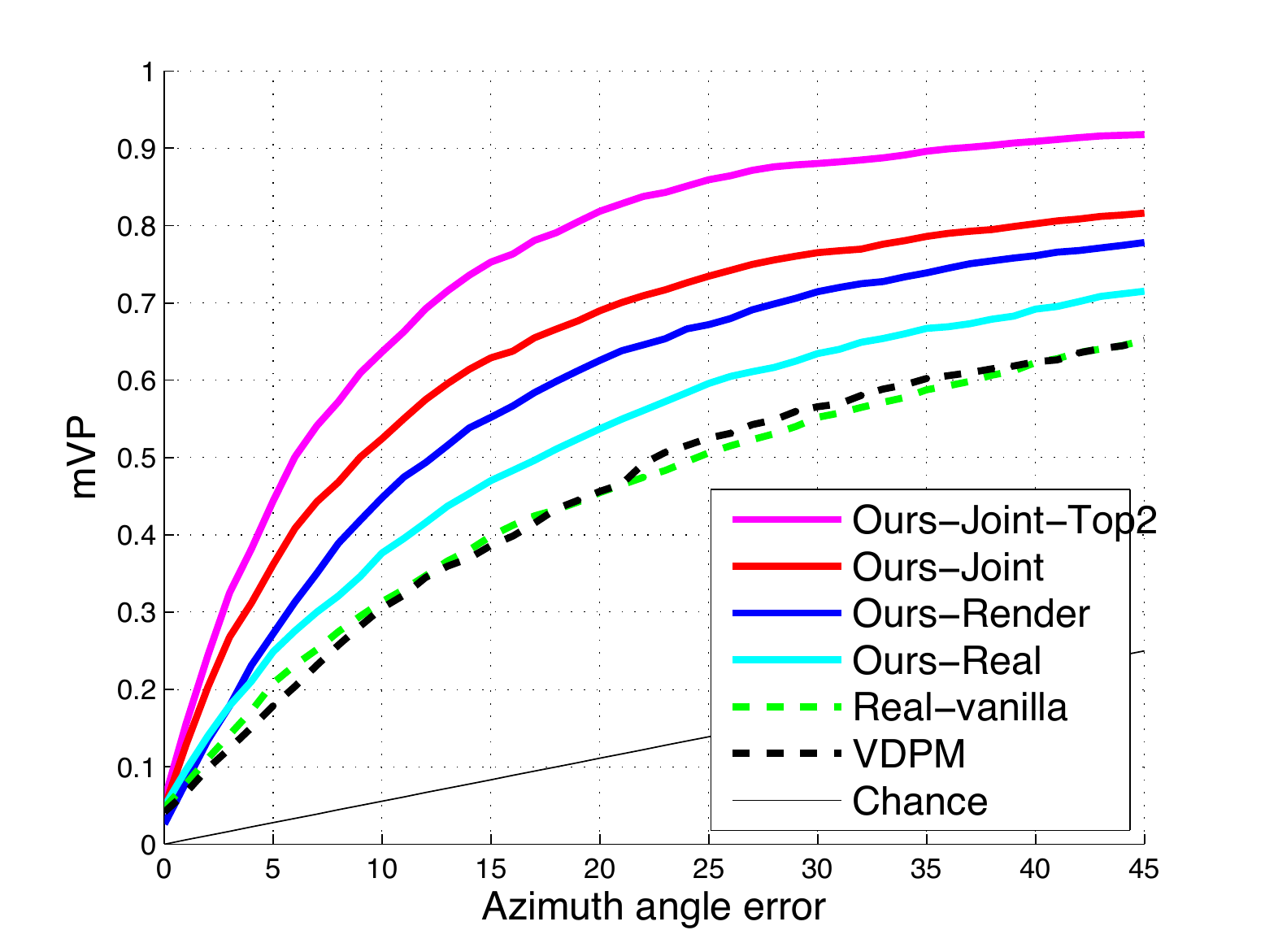}
    \end{subfigure}
    \begin{subfigure}
        \centering
        \includegraphics[width=0.48\linewidth]{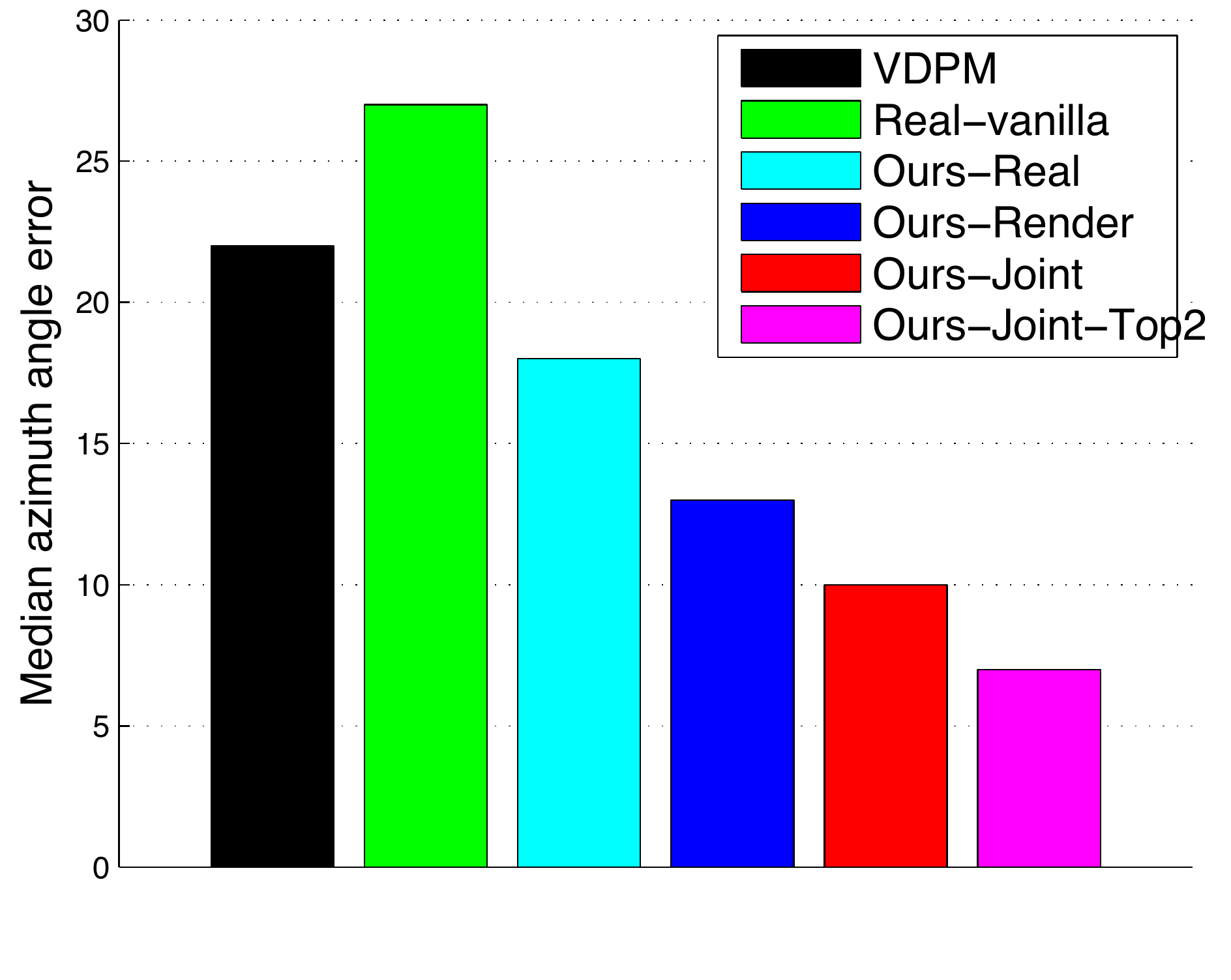}
    \end{subfigure}
    \caption{{\bf Viewpoint estimation performance using positive VDPM detection windows on PASCAL VOC 2012 val.} \emph{Left:} mean viewpoint estimation accuracy as a function of azimuth angle error $\delta_{\theta}$. Viewpoint is correct if distance in degree between prediction and groundtruth is less than $\delta_{\theta}$. \emph{Right:} medians of azimuth estimation errors (in degree), the lower the better.  Refer to Table~\ref{tb:settings} for settings of methods in comparison.
    }
    \label{fig:vp_append}
\end{figure}

\section{Comparison over VDPM for Viewpoint Estimation by VDPM Bounding Box (Sec 5.2)}
\label{sec:vdpm_comparison}
In Figure~\ref{fig:joint_detection_and_view_estimation} of the main paper, we compare viewpoint estimation performance of our methods versus VDPM, using detection windows from R-CNN. Here, we compare these methods again, using detection windows from VDPM, i.e., to estimate the viewpoint of objects in detection windows from VDPM. To make this material more self-contained, we summarize the settings of all methods in Table~\ref{tb:settings}. 
\label{sec:comparison_over_VDPM}

Results are shown in Figure~\ref{fig:vp}. The trend is unchanged, except that in Figure~\ref{fig:vp} VDPM (16V) is slightly better than Real-vanilla (model trained with real images, without using our new loss function). Note that R-CNN detects many more difficult cases than VDPM in terms of occlusion and truncation. In other words, Figure~\ref{fig:vp} focuses on the comparison of our methods and VDPM over simple cases.

\section{Quantitative Results on elevation and in-plane rotation (Sec 5.5)}
\label{sec:phi_tau}
\begin{figure}[h!]
    \begin{subfigure}
        \centering
        \includegraphics[width=0.48\linewidth]{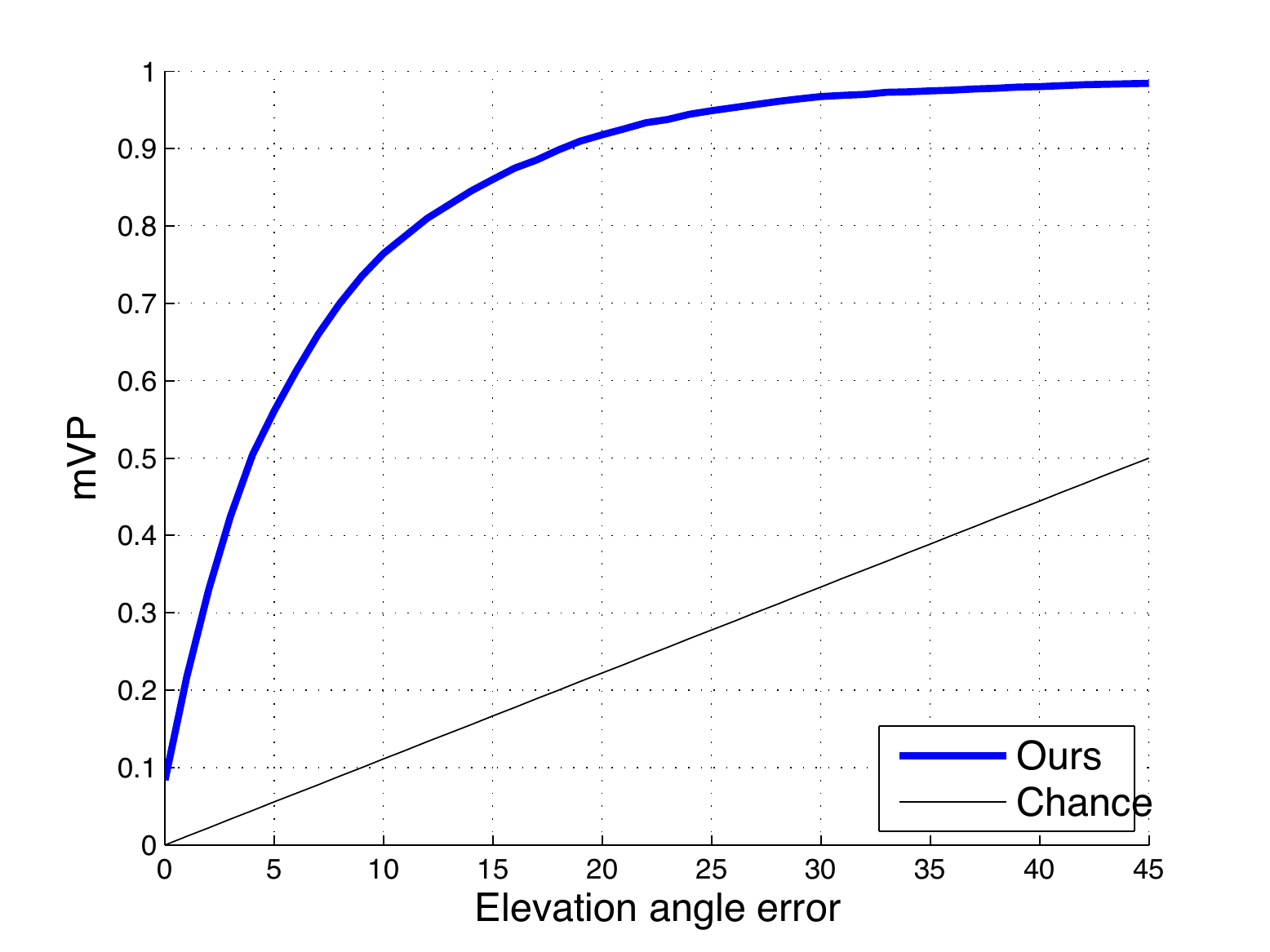}
    \end{subfigure}
    \begin{subfigure}
        \centering
        \includegraphics[width=0.48\linewidth]{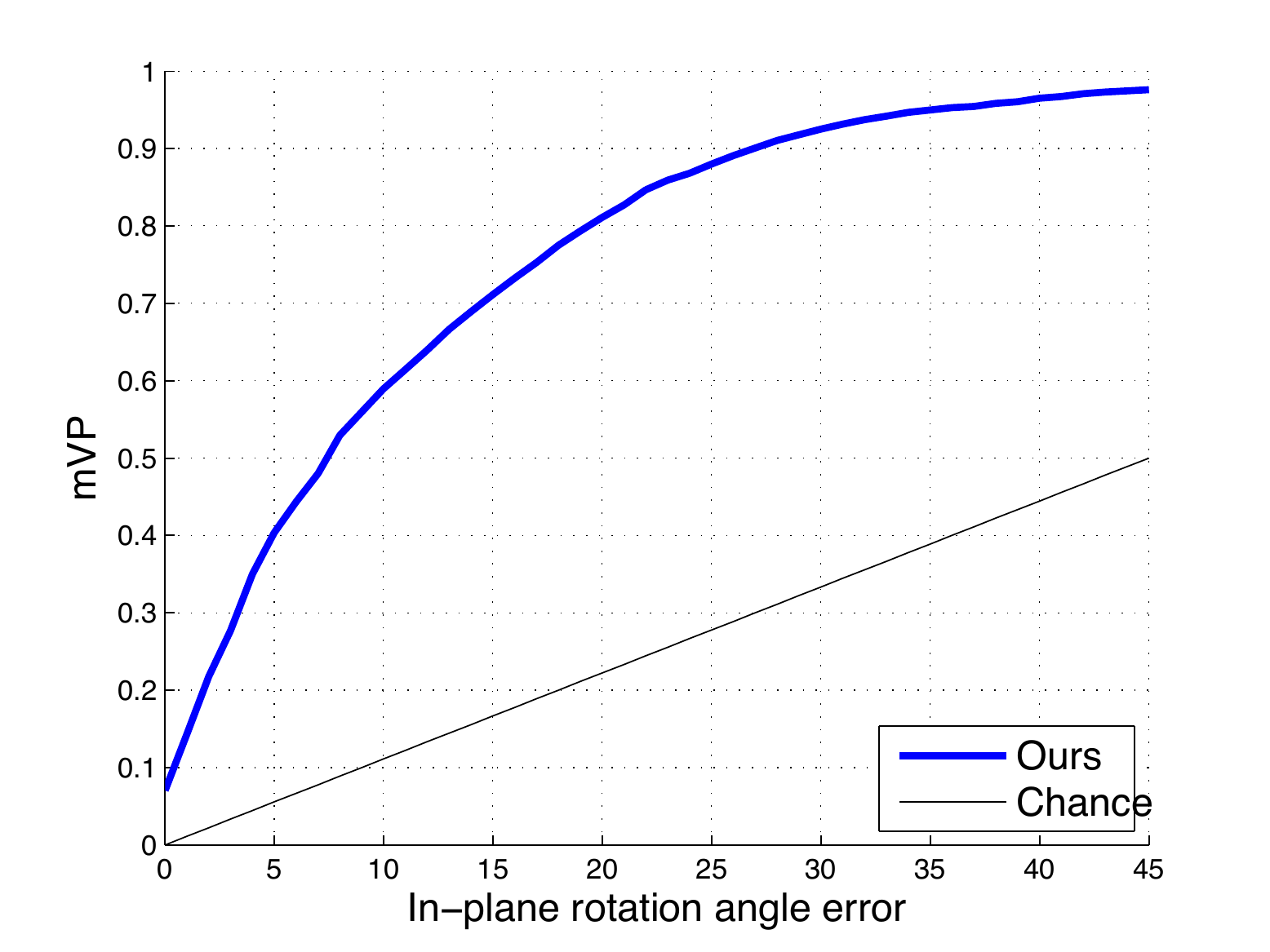}
    \end{subfigure}
    \caption{{\bf Viewpoint estimation precision on elevation and in-plane rotation.} We show test results on VOC val in PASCAL3D+. Our model here is trained with rendered images only. \emph{Left:} Viewpoint precision for elevation. \emph{Right:} Viewpoint precision for in-plane rotation. }
    \label{fig:elevation_tilt}
\end{figure}

In Figure~\ref{fig:elevation_tilt}, we show results on elevation and in-plane rotation estimation. Since most objects tend to have small elevation and in-plane rotation variations, the range of those two parameters are smaller compared with azimuth angle, thus the viewpoint estimation accuracy is also higher.

\section{Synthesis Pipeline Details (Sec 4.1)}
\label{sec:synthesis_pipeline}

\begin{figure*}[t!]
\centering
	\includegraphics[width=0.80\linewidth]{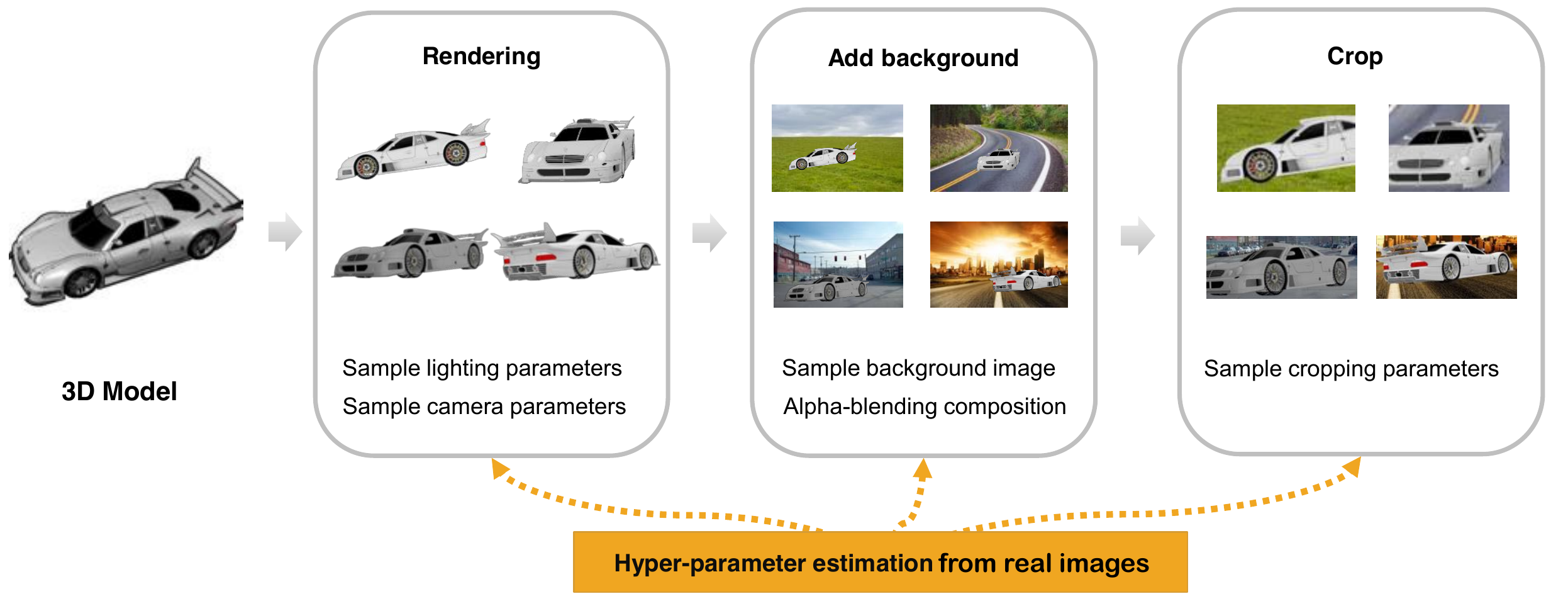}
	\caption{{\bf Scalable and overfit-resistant synthesis pipeline.} }
	\label{fig:synthesis_pipeline}
\end{figure*}
The synthesis pipeline is illustrated in Figure~\ref{fig:synthesis_pipeline}.

\paragraph{3D Model Normalization} Following the convention of PASCAL 3D+ annotation, all 3D models are normalized to have the bounding cube centered at the origin and diagonal length $1$. 

\paragraph{Rendering} The parameters that affect the rendering include  lighting condition, camera extrinsics and intrincis. To render each image, we sample a set of these parameters from a distribution as follows:

\begin{itemize}
\item {\bf Lighting condition}. We add $N$ point lights and enable the environmental light. $N$ is uniformly sampled from $1$ to $10$. All lighting parameters are sampled i.i.d. The position $p_{light}$ is uniformly sampled on a sphere of radius $14.14$, between latitude $0\deg$ and $60\deg$; the energy $E \sim \mathcal{N}(4,3)$; the color is fixed to be white. 

\item {\bf Camera extrinsics.} 
We first describe the camera position parameters (i.e., translation parameters). In the polar coordinate system, let $\rho$ be the distance of the optical center to the origin, $\theta$ and $\phi$ be the longitude and latitude, respectively. We use kernel density estimation (KDE) to estimate the non-parametric distributions of $\rho, \theta, \phi$ for each category respectively, from the VOC 12 train set of PASCAL3D+ dataset. We then use the estimated distributions to generate i.i.d. samples for rendering. 

We then describe the camera pose parameters (i.e., rotation parameters). Similar to the position parameters, we use KDE to estimate distribution of in-plane rotation $\psi$ for each category and generate i.i.d. samples for rendering. We set the camera to look at the origin, the image plane perpendicular to the ray from the optical center to the origin.

\item {\bf Camera intrinsics.} The focal length and aspect ratio are fixed to be $35$ and $1.0$, respectively.
\end{itemize}

We use Blender, an open-source 3D graphics and animation software, to efficiently render each 3D model. 

\paragraph{Background synthesis} All details are described in the main paper.

\paragraph{Cropping} 
\begin{figure}[h!]
\centering
	\includegraphics[width=\linewidth]{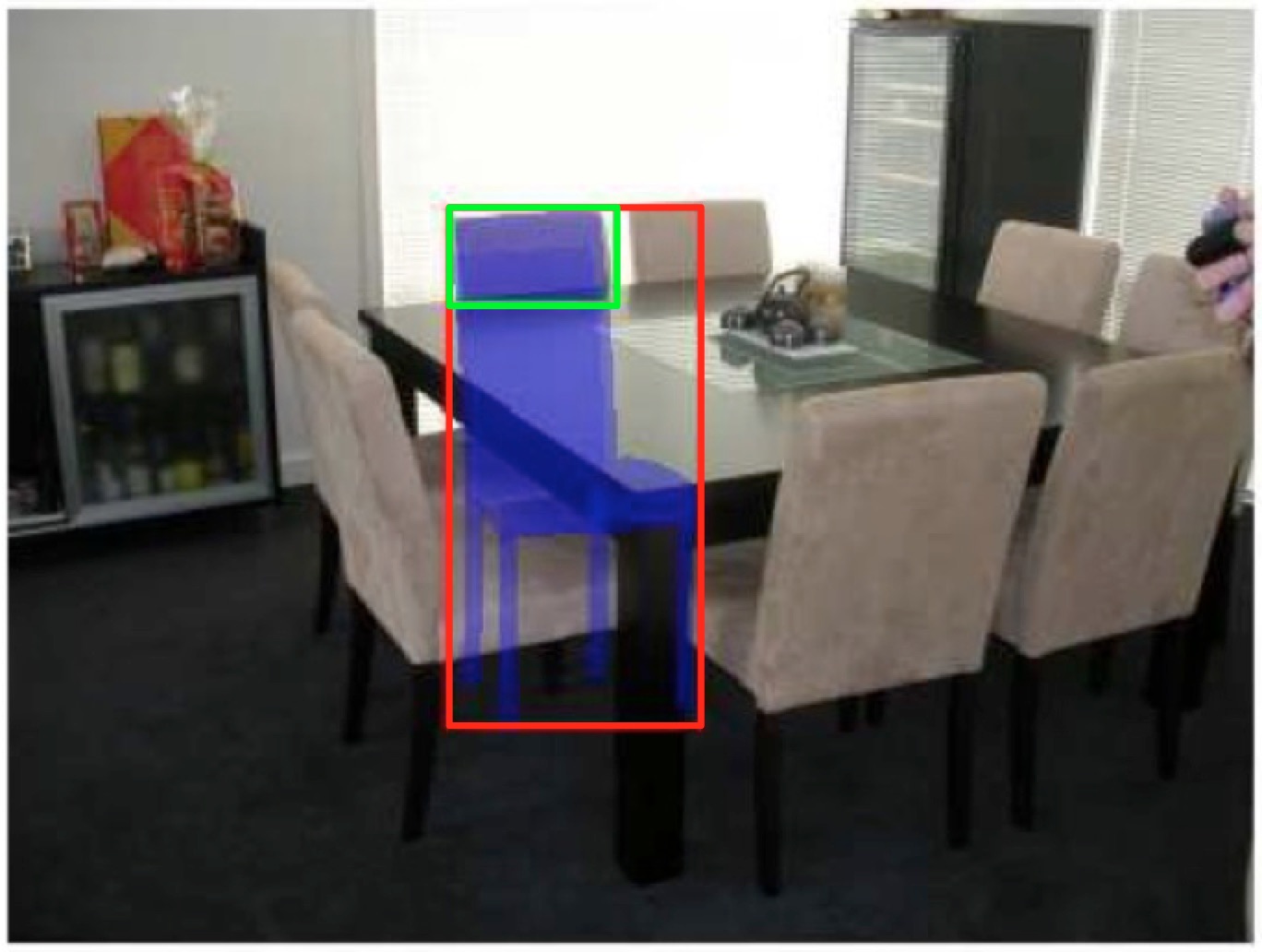}
	\caption{{\bf Cropping parameter estimation.} We estimate the cropping parameters by comparing the groundtruth bounding box and full object bounding box. The groundtruth bounding box (green) is from PASCAL 3D+. The full object bounding box (red) is estimated by us as follows: because the real training image dataset (PASCAL 3D+) has provided landmark registrations between each object instance and a similar 3D model, we can project the 3D model to the image space and estimate the full bounding box.}
	\label{fig:cropping}
\end{figure}

As seen in Figure~\ref{fig:cropping}, we use annotations provided by PASCAL3D+ to get truncation patterns of objects in real images. For each real training image, we project 3D model corresponding to the object back to the image and then get a bounding box for the projected model (full object bounding box). Then, by comparing with the provided groundtruth bounding box of the object, we know how the object is truncated. Specifically, we know the relative position of four edges between full box and groundtruth box. We use kernel density estimation to learn non-parametric distribution of these relative positions for each category and generate samples for croppings of rendered images.

\section{Network Details (Sec 4.2)}
\label{sec:network_details}
We adapt network architecture from R-CNN to object viewpoint estimation. For notation, conv means convolutional layer including pooling and ReLu. fc means fully connected layer. The number following conv or fc, starting from 1 as bottom, means order of layer. We keep structures of convolutional layers and fc6 and fc7 fully connected layers consistent with R-CNN network.   Depending on the number of layers fine tuned (for example, we fine tune layers above, not including, conv3), we can use shared lower layers for both detection and viewpoint estimation, which reduces computation cost. The last fully connected layer is object category specific. All categories share layers below and including fc7 (trained by viewpoint annotations, fc7 features now preserve geometric information about the image). On top of fc7, each category has a fully connected layer with 8640 neurons (4320 for azimuth + 2160 for elevation + 2160 for in-plane rotation). We use the geometric structure aware loss mentioned in the main paper as the loss layer. During back propagation, only viewpoint losses from the object category of the instance will be counted. 
\begin{table}[h!]
\centering
\begin{tabular}{l|c|l}
	\hline
	Name & synset offset & num\\
	\hline
    aeroplane & n02691156 & 4045\\
    bicycle & n02834778 & 59\\
    boat & n04530566 & 1939\\
    bottle & n02876657 & 498\\
    bus & n02924116 & 939\\
    car & n02958343 & 7497\\
	chair & n03001627 & 6928\\
	dining table & n04379243 & 3650\\
	motorbike & n03790512 & 337\\
	sofa & n04256520 & 3173\\
	train & n04468005 & 389 \\
	tv monitor & n03211117& 1095\\
	\hline
\end{tabular}
\caption{Statistics of models used in the paper. }
\label{tb:model_statistics}
\end{table}

\section{More Details on 3D Model Set (Sec 5.1)}
\label{sec:model_statistics}
Table~\ref{tb:model_statistics} lists the statistics of the models used in the main paper. All models are downloaded from ShapeNet, which is organized by the taxonomy of WordNet. Models are also pre-aligned to have consistent orientation by ShapeNet. In WordNet (and ShapeNet), each category is indexed by a unique id named ``synset offset''.

\section{More Examples}
\label{sec:examples}
See next pages for examples of positive and negative results. The negative results are grouped by error patterns.
\begin{figure*}
	\centering
	\includegraphics[width=\textwidth]{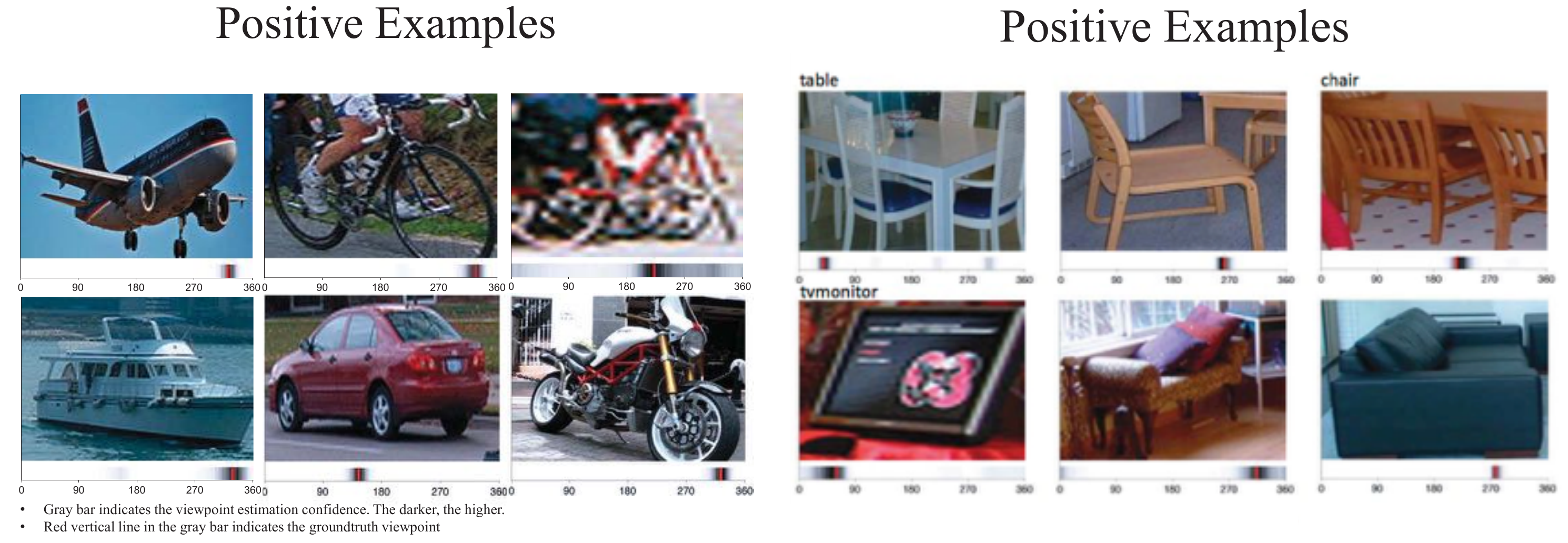}
    ~\\
	\includegraphics[width=0.45\linewidth]{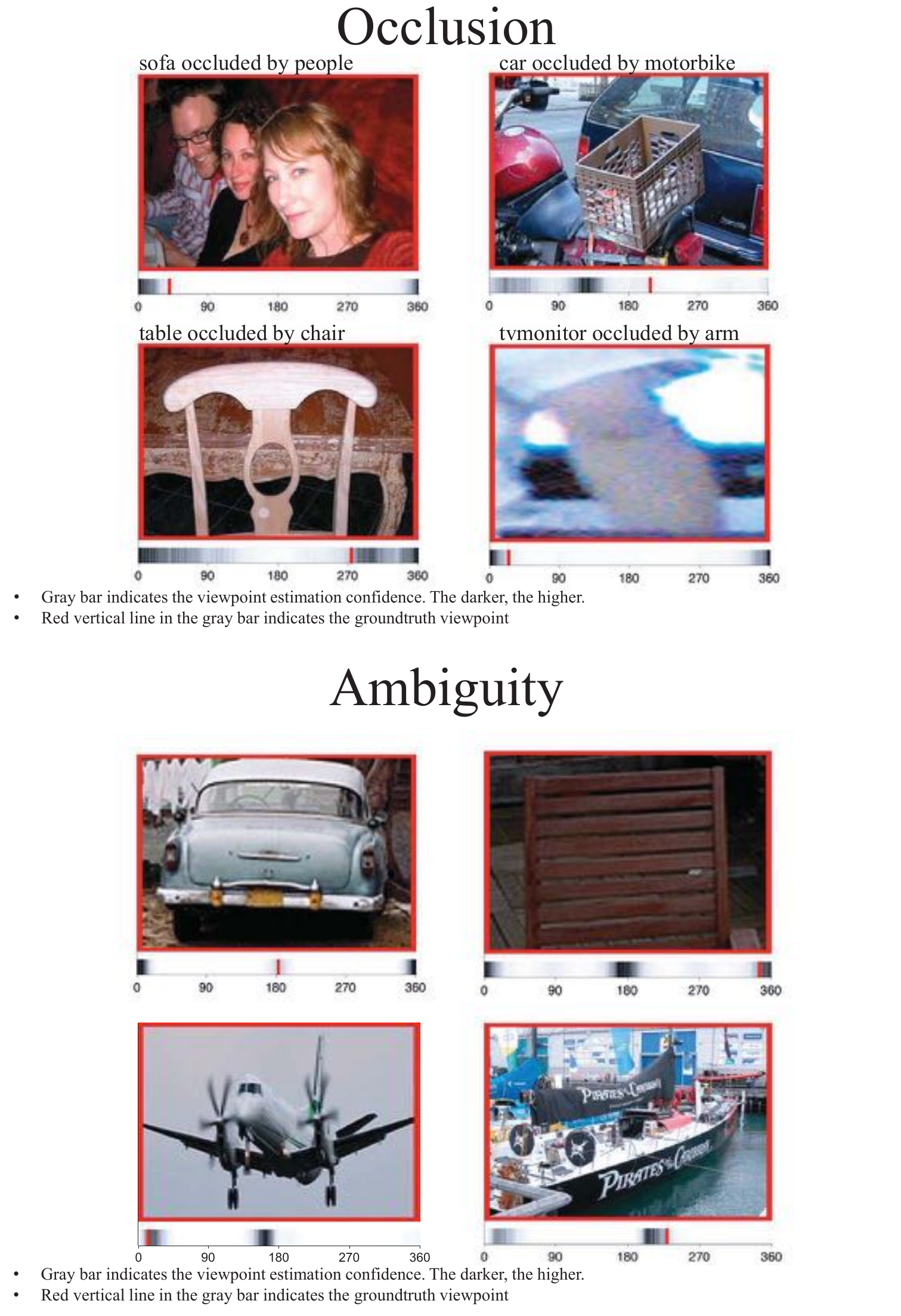}
	\qquad\includegraphics[width=0.45\linewidth]{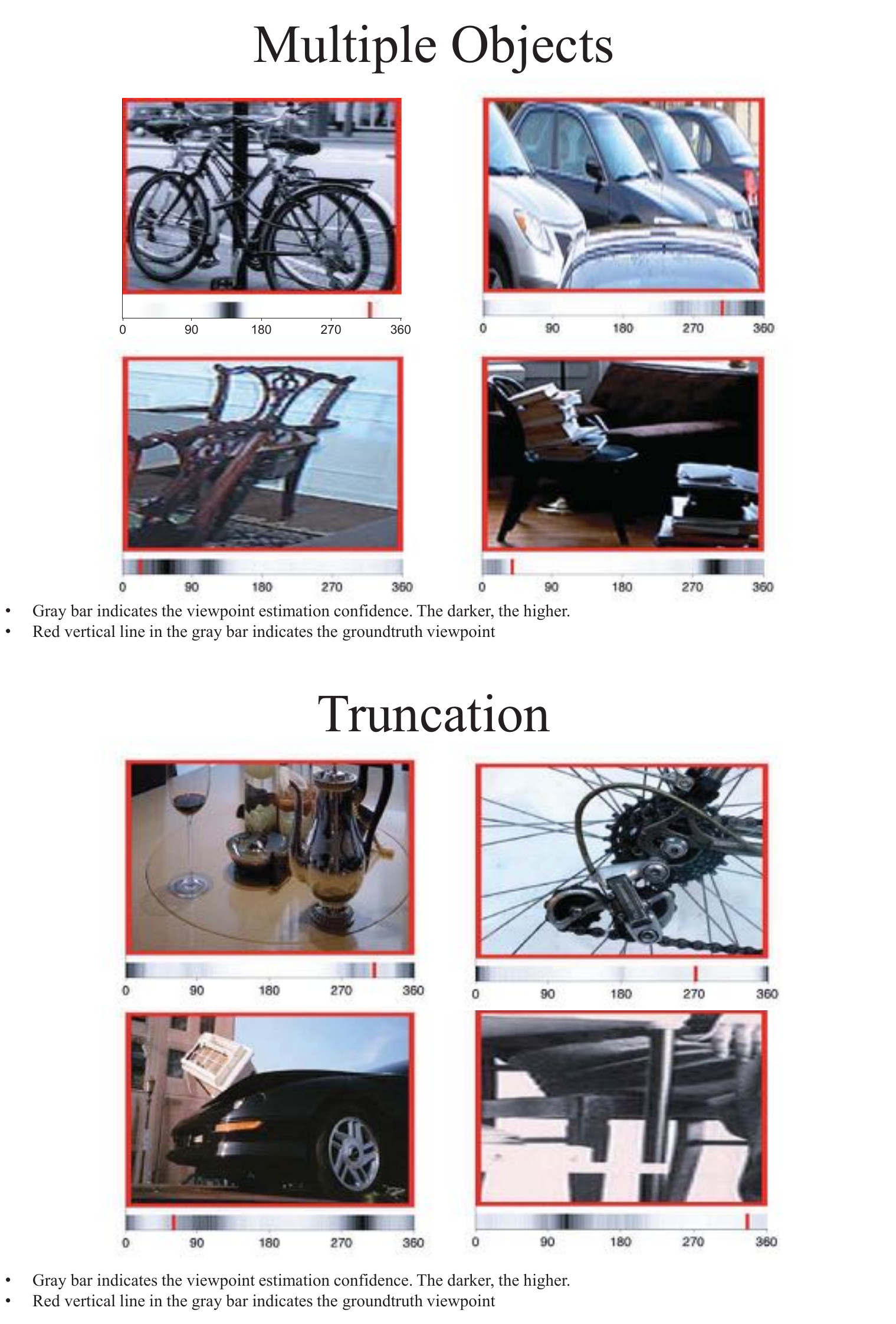}
\end{figure*}

\end{document}